\documentclass[review]{elsarticle}

\usepackage{lineno,hyperref}
\usepackage{xcolor,hhline}
\usepackage{bm}
\usepackage{multirow}
\usepackage{color, colortbl, amsmath}
\usepackage[linesnumbered,ruled,noend]{algorithm2e}
\usepackage{cleveref}[2012/02/15]
\crefformat{footnote}{#2\footnotemark[#1]#3}

\journal{Knowledge-based Systems}

\begin{document}

\begin{frontmatter}

\title{Handling Imbalanced Datasets Through Optimum-Path Forest}

\author[1]{Leandro Aparecido Passos$^\ast$}	
\author[1]{Danilo S. Jodas$^\ast$}
\author[1]{Luiz C. F. Ribeiro$^\ast$}
\author[2]{Marco~Akio}
\author[2]{Andre Nunes de Souza}
\author[1]{Jo\~{a}o Paulo Papa}

\address[1]{Department of Computing, S\~ao Paulo State University\\
	Av. Eng. Luiz Edmundo Carrijo Coube, 14-01, Bauru, 17033-360, Brazil\\
	email:\{leandro.passos, luiz.felix, joao.papa\}@unesp.br, danilojodas@gmail.com}

\address[2]{Department of Electrical Engineering, S\~ao Paulo State University\\
	Av. Eng. Luiz Edmundo Carrijo Coube, 14-01, Bauru, 17033-360, Brazil\\
	email: ikeshoji@uol.com.br, andre.souza@unesp.br\\
$^\ast$Authors contributed equally.}

\begin{abstract}
In the last decade, machine learning-based approaches became capable of performing a wide range of complex tasks sometimes better than humans, demanding a fraction of the time. Such an advance is partially due to the exponential growth in the amount of data available, which makes it possible to extract trustworthy real-world information from them. However, such data is generally imbalanced since some phenomena are more likely than others. Such a behavior yields considerable influence on the machine learning model's performance since it becomes biased on the more frequent data it receives. Despite the considerable amount of machine learning methods, a graph-based approach has attracted considerable notoriety due to the outstanding performance over many applications, i.e., the Optimum-Path Forest (OPF). In this paper, we propose three OPF-based strategies to deal with the imbalance problem: the $\text{O}^2$PF and the OPF-US, which are novel approaches for oversampling and undersampling, respectively, as well as a hybrid strategy combining both approaches. The paper also introduces a set of variants concerning the strategies mentioned above. Results compared against several state-of-the-art techniques over public and private datasets confirm the robustness of the proposed approaches.
\end{abstract}


\end{frontmatter}


\section{Introduction}\label{s.intro}

Imbalanced class distribution in a dataset represents a major concern for the effectiveness of many machine learning models since they might be prone to assign new samples to the majority class at the detriment of the minority one. Current baseline strategies adopted to tackle the imbalance issue include: (i) improve the classification model to make it more robust regarding minority class \cite{o2019random, chen2004using}, and (ii) change of the dataset distribution to match the number of samples in each class. Regarding the latter strategy, the balancing of the dataset consists of either reducing the samples of the majority class or including new synthetic samples to the minority class, denoting the task of undersampling and oversampling, respectively.

\begin{sloppypar}
Many studies addressed oversampling techniques to tackle imbalanced datasets in practical applications \cite{wang2021ponzi,jiang2021windturbine,sleeman2021bigdata,OksuzPAMI:2020,JingPAMI:19,HuangPAMI:19,KovacsASC:2019, MalhotraNC:2019, cordon2018imbalance, saez2016analyzing}, including the well-known Synthetic Minority Over-sampling Technique (SMOTE)~\cite{ChawlaJAIR:02} and its variations borderline-SMOTE1 and borderline-SMOTE2~\cite{han:05borderline}, the Adaptive Synthetic (ADASYN)~\cite{He:2008adasyn}, the Combined Synthetic Oversampling and Undersampling Technique (CSMOUTE), the LR-SMOTE~\cite{liang2020lrsmote}, and the Class Decomposition SMOTE (CDSMOTE)~\cite{elyan2021cdsmote}, to cite a few. Although widely used to handle the class imbalance in machine learning problems, the major shortcoming of SMOTE is the introduction of noise data in the majority class samples since the labels among all classes are not considered in the interpolation process. Furthermore, several clustering-based oversampling techniques exist in the literature, with particular reference to ones such as Majority Weighted Minority Oversampling Technique (MWMOTE)~\cite{Barua2014MWMOTE}, $k$-means SMOTE~\cite{Douzas2018KmeansSmote}, Self-Organizing Map Oversampling (SOMO)~\cite{douzas2017self} and Agglomerative Hierarchical Clustering (AHC)~\cite{Cohen2006AHC}.
\end{sloppypar}

\begin{sloppypar}
Considering the undersampling strategies, one can mention the nearest neighbor-based approaches, in which the majority class samples are kept or removed by considering their nearest samples of the minority class. NearMiss family methods~\cite{maniWLID2003} and the Condensed Nearest Neighbor (CNN)~\cite{hart1968condensed} are examples of undersampling approaches based on nearest neighbors. However, the main drawback of Near Miss methods regards the holding of majority samples only in cases where the number of minority class neighbors is large. Condensed Nearest Neighbor, although proposed to reduce the computational cost of the K-Nearest Neighbor (KNN) method, still suffers from outliers and the overlapping among classes. In a similar manner, Koziarski~\cite{koziarski2020radial} introduced a Radial-Based Undersampling (RBU) approach, obtaining promising results. Further, Cervellera and Macci\`{o}~\cite{cervellera2020voronoi} introduced a Voronoi diagram-based approach to perform either under or oversampling through a Gaussian Mixture Model.
\end{sloppypar}

To deal with the issues mentioned above for both oversampling and undersampling tasks, this paper proposes a novel framework extending the Optimum Path Forest (OPF) algorithm capabilities to handle the problem of imbalanced datasets. The motivation for such an approach lies in the competence of the OPF attributes to tackle a wide variety of problems, such as supervised classification~\cite{PapaIJIST:09,PapaPR:2012,PapaPRL:17}, clustering~\cite{RochaIJIST:09}, and anomaly detection~\cite{guimaraesIEEENETWORK:2018}. Further, it has been successfully employed to solve problems concerning intricate data distribution~\cite{SouzaIEEETFS:19}, and to improve other techniques, such as the radial basis function neural networks~\cite{rosaICPR:2014} and the brainstorm optimization algorithm~\cite{afonsoSACI:2018}, to cite a few.

Regarding the oversampling task, Passos et al.~\cite{passosSCBMS:2020} proposed the Oversampling Optimum Path Forest ($\text{{O}}^2$PF), an unsupervised OPF-based approach that captures intrinsic characteristics of minority class samples through clustering approaches, and afterward generates new synthetic samples using a Gaussian distribution over the mean and the covariance matrix of each cluster. This last step is conceived to guarantee the variability of the new synthetic samples while enforcing them to share similar characteristics with the real ones. Moreover, this paper introduces four variations of the method, considering different policies for generating new samples.

Concerning the undersampling approach, the proposed OPF-based framework for data imbalance introduces the Optimum Path Forest-based approach for undersampling, the so-called OPF-US, which concerns in removing samples that are more likely to lead the classifier to ill decision boundaries. To such an extent, the supervised OPF is employed to capture the importance of the majority class samples via the k-fold cross-validation approach considering five splits of the training set. After the training phase, a rank score representing the prediction of a validation sample is assigned to each training sample in a k-fold cross-validation iteration. At the final stage, the training samples are ordered by the ascending order of their average rank scores, being the majority class samples with lower ranks removed from the training set until it is fully balanced. Furthermore, this work also introduces three variants of OPF-US, which adopt different policies regarding the samples to be pruned. Besides, it also proposes three hybrid approaches considering the tasks of data undersampling and oversampling.

Different from the approaches mentioned above, whose primary focus consists of aggregating variants of the SMOTE into existing clustering algorithms, or even using existing clustering techniques, this paper shows a different approach to explore the potential of a fully OPF-based method for oversampling and undersampling tasks. Therefore, the main contributions of this paper are fourfold, listed as follows:

\begin{itemize}
    \item to propose four variations of the oversampling algorithm $\text{O}^2$PF, i.e., $\text{O}^2$PF Radius Interpolation, $\text{O}^2$PF Mean Interpolation, $\text{O}^2$PF Prototype, and $\text{O}^2$PF Weight Interpolation;
    \item to propose four OPF-based methods for data undersampling, i.e.,  OPF-US, OPF-US1, OPF-US2, and OPF-US3;
    \item to propose three hybrid approaches combining both OPF-based methods for undersampling and oversampling, i.e., OPF-US1-$\text{O}^2$PF, OPF-US2-$\text{O}^2$PF, and OPF-US3-$\text{O}^2$PF;
    \item to foster the literature regarding methods to deal with imbalanced datasets, since it establishes new state-of-the-art methods to tackle the problem.
\end{itemize}

This paper is organized as follows: Section~\ref{s.theoretical} presents the theoretical basis of the Optimum Path Forest algorithm for supervised and unsupervised classification purposes. Further, Section~\ref{s.approaches} describes the theory behind the undersampling and oversampling of imbalanced datasets via OPF. Sections~\ref{s.methodology} and~\ref{s.experiments} provide the methodology and the experimental results, respectively. Finally, Section~\ref{s.conclusions} states the conclusions and future work.

\section{Theoretical Background}
\label{s.theoretical}

This section introduces the main concepts regarding the supervised and unsupervised Optimum-Path Forest.

\subsection{Supervised Optimum-Path Forest} 
\label{ss.supOPF}

The supervised version of the Optimum-path Forest~\cite{PapaIJIST:09,PapaPR:2012} can be defined as a graph-based classifier that assumes each data sample as a node, and the connection between each pair of samples as an edge in the graph. Additionally, the model selects the most representative samples, i.e., prototypes, to compete among themselves in order to provide optimum-path costs to the remaining samples in a conquering-like process. Therefore, the training process is accomplished by the minimization of the path-cost function $f_{max}$, as follows:

\begin{eqnarray}
 \label{e.fmax}
f_{max}(\langle
\bm{q}\rangle) & = & \left\{ \begin{array}{ll}
  0 & \mbox{if $\bm{q}\in {\cal P}$,} \\
  +\infty & \mbox{otherwise}
  \end{array}\right. \nonumber \\
  f_{max}(\phi_{\bm{q}} \cdot \langle \bm{q},\bm{u} \rangle) & = & \max\{f_{max}(\phi_{\bm{q}}),d(\bm{q},\bm{u})\}, 
\end{eqnarray}
where ${\cal P}$ denotes the set prototypes, $\phi_{\bm{q}}$ denotes a sequence of adjacent samples, i.e., a path, starting from a root in ${\cal P}$ and with terminus at sample ${\bm{q}}$, and $d(\bm{q},\bm{u})$ represents the distance between samples $\bm{q}$ and $\bm{u}$. Moreover, $\phi_{\bm{q}} \cdot \langle \bm{q},\bm{u} \rangle$ stands for the concatenation between the path $\phi_{\bm{q}}$ and the edge $\langle \bm{q},\bm{u} \rangle$. In a nutshell, $f_{max}(\phi_{\bm{q}})$ stands for the maximum distance among adjacent samples in the path $\phi_{\bm{q}}$.

Let ${\cal P}^\ast\subseteq{\cal P}$ be the set of prototypes that minimize the error during training, which can be found by computing the Minimum Spanning Tree over the training set and selecting adjacent samples with different labels. The training step is performed by assigning an optimum cost $C(\bm{u})$ to each sample $\bm{u} \in {\cal V}$, where ${\cal V}$ stands for the training set, i.e.:

\begin{equation}
\label{e.conquering_function}
	C(\bm{u})  =  \min_{\forall \bm{q} \in {\cal V}}\{\max\{C(\bm{q}),d(\bm{q},\bm{u})\}\}, 
\end{equation}
where $\bm{q}$ stands for the training sample that conquered $\bm{u}$. The classification step is performed by finding out the training sample that offers the optimum-path cost to each test sample in a similar fashion to Equation~\ref{e.conquering_function}. Algorithm~\ref{a.opf_sup} implements the training procedure mentioned above.

\IncMargin{1em}
\begin{algorithm}[!h]
\SetKwInput{KwInput}{Input}
\SetKwInput{KwOutput}{Output}
\SetKwInput{KwAuxiliary}{Auxiliary}

\caption{Supervised OPF training algorithm}
\label{a.opf_sup}
\Indm
\KwInput{A training set ${\cal V}$, set of prototypes ${\cal P}^\ast\subseteq {\cal V}$, map of training set labels $\lambda$,  and distance function $d$.}
\KwOutput{Predecessor map $O$, path-cost map $C$, and label map $L$.}
\KwAuxiliary{Priority queue $Q$, and variable $cst$.}
\Indp\Indpp

\BlankLine
\For{all $\bm{q} \in {\cal V}$}{
	$O(\bm{q}) \leftarrow nil$, $C(\bm{q}) \leftarrow +\infty$\;
}

\For{all $\bm{q} \in {\cal P}^\ast$}{
	$C(\bm{q}) \leftarrow 0$, $L(\bm{q}) = \lambda(\bm{q})$, $Q \leftarrow \bm{q}$\;
}

\While{$Q \neq \emptyset$}{
	Remove from $Q$ a sample $\bm{q}$ such that $C(\bm{q})$ is minimum\;

	\For {each sample $\bm{u} \in {\cal V}$ such that $\bm{q} \neq \bm{u}$ and $C(\bm{u}) > C(\bm{q})$}{
		$cst \leftarrow \max\{C(\bm{q}), d(\bm{q},\bm{u})\}$\;
		\If{$cst < C(\bm{u})$}{
			\lIf{$C(\bm{u}) \neq +\infty$}{Remove $\bm{u}$ from $Q$}
			$L(\bm{u}) \leftarrow L(\bm{q})$, $O(\bm{u}) \leftarrow \bm{q}$, $C(\bm{u}) \leftarrow cst$, $Q \leftarrow \bm{u}$\;
		}
	}
}
\Return{$[O, C, L]$}
\end{algorithm}
\DecMargin{1em}

Lines $1-4$ initialize the cost map by assigning zero cost to the prototypes (Lines $3-4$) and a high cost to the remaining nodes (Lines $1-2$). Besides, all samples have their predecessors set to $nil$, and the prototypes are inserted into the priority queue $Q$. Notice the main loop in Lines $5-11$ concerns the core of the OPF algorithm. A sample with minimum cost is removed from $Q$, and its neighborhood is analyzed: if the cost $cst$ offered to a neighboring sample $\bm{u}$ is lower than its current cost, it receives the label assigned to $\bm{q}$, and it is further added to its optimum-path tree (Line $11$). Groups of samples from the same class might be represented by multiple optimum-path trees, and there must be at least one per class.

Considering the implementation provided in Algorithm~\ref{a.opf_sup}, Lines $1-4$ take $(\theta(\left|{\cal V}\right|)+\theta(\left|{\cal P}^\ast\right|))\in\theta(\left|{\cal V}\right|)$ steps. In Lines $5-11$, $\theta(\left|{\cal V}\right|)$ training samples are added to the priority queue $Q$, each one removed once. Therefore, the loop is executed $\theta(\left|{\cal V}\right|)$ times. Additionally, once the model employs a binary heap to implement the priority queue, Line $6$ has complexity of $O(\log\left|{\cal V}\right|)$. Moreover, considering the process is performed over a complete graph, the inner loop presented in Lines $7-11$ runs $\theta(\left|{\cal V}\right|)$ times. Thereinafter, the final complexity for training is $\theta(\left|{\cal V}\right|^2)$. The classification step can be performed in $\theta(\left|{\cal V}\right|.\left|{\cal J}\right|)$, where $\left|{\cal J}\right|$ represents the test set size. However, Papa et al.~\cite{PapaPR:2012} showed it can be performed in $O(\left|{\cal V}\right|.\left|{\cal J}\right|)$.

\subsection{Unsupervised Optimum-Path Forest} 
\label{ss.unsupOPF}

In its unsupervised version, the Optimum-Path Forest performs clustering by representing each training sample as a graph node. Such samples are connected to their $k$-nearest neighbors and the arcs are weighted according to the distance between them. Further, each node is weighted by a probability density function (pdf), as follows:

\begin{eqnarray}
  \rho(\bm{q}) & = & \frac{1}{\sqrt{2\pi\psi^2}k} \sum_{\forall \bm{u}\in {\cal A}_k(\bm{q})} \exp\left(\frac{-d^2(\bm{q},\bm{u})}{2\psi^2}\right), \label{e.density}
\end{eqnarray}
where ${\cal A}_k(\bm{q})$ denotes the $k$-neighborhood of sample $\bm{q}$, $\psi = \frac{m_w}{3}$, and $m_w$ is the maximum weight among all edges in the graph.

\begin{sloppypar}
The probability density is estimated similarly to a Parzen-window. Such a procedure requires discovering the optimum number of nearest neighbors ${k^{\ast} \in \{1 \leq  k_{\max}\leq |{\cal V}|\}}$, where $k_{max}$ denotes a hyperparameter that is commonly computed by finding $k^{\ast}$ that minimizes the graph cut over ${\cal V}$.	
\end{sloppypar}

The model establishes a set of prototypes ${\cal P}$ composed of one element per maximum of the pdf, and $\bm{u}$ is assigned to the path whose minimum density value along it is maximum, i.e.:

\begin{eqnarray}
\label{e.fmin}
f_{min}(\langle \bm{u} \rangle) & = & \left\{ \begin{array}{ll} 
    \rho(\bm{u})           & \mbox{if $\bm{u} \in {\cal P}$} \\
    \rho(\bm{u}) - \delta  & \mbox{otherwise,}
 \end{array}\right. \nonumber \\
f_{min}(\langle \phi_{\bm{q}}\cdot \langle \bm{q},\bm{u}\rangle\rangle)&=& \min \{f_{min}(\phi_{\bm{q}}), \rho(\bm{u})\},
\end{eqnarray}
where $\delta$ is small constant. Algorithm~\ref{a.unsupopf} implements the unsupervised Optimum-Path Forest algorithm.

\IncMargin{1em}
\begin{algorithm}[!h]
\SetKwInput{KwInput}{Input}
\SetKwInput{KwOutput}{Output}
\SetKwInput{KwAuxiliary}{Auxiliary}

\caption{OPF Clustering algorithm}
\label{a.unsupopf}
\Indm
\KwInput{Graph ${\cal G} = ({\cal V},{\cal A}_k)$.}
\KwOutput{Predecessor map $O$, path-cost map $C$, and cluster label map $L$.}
\KwAuxiliary{Priority $Q$, and variables $cst$ and $l\leftarrow 1$.}
\Indp\Indpp

\BlankLine
\For{all $\bm{q} \in {\cal V}$}{
	Compute $\rho(\bm{q})$ using Equation~\ref{e.density}\;
	$O(\bm{q})\leftarrow nil$, $C(\bm{q})\leftarrow \rho(\bm{q})-\delta$, $Q\leftarrow \bm{q}$\;
}

\While{$Q \neq \emptyset$}{
	Remove from $Q$ a sample $\bm{q}$ such that $C(\bm{q})$ is maximum\;
	\If{$O(\bm{q}) = nil$}{
		$L(\bm{q})\leftarrow l$, $l\leftarrow l + 1$, and $C(\bm{q})\leftarrow \rho(\bm{q})$\;
	}
	\For {all $\bm{u}\in {\cal A}_k(\bm{q})$ such that $C(\bm{u}) < C(\bm{q})$}{
	$cst\leftarrow \min\{C(\bm{q}), \rho(\bm{u})\}$\;
		\If{$ cst > C(\bm{u})$}{
			$L(\bm{u})\leftarrow L(\bm{q})$, $O(\bm{u})\leftarrow \bm{q}$, $C(\bm{u})\leftarrow cst$\;
			Update position of $\bm{u}$ in $Q$\;
		}
	}
}
\Return $[O,C,L]$
\end{algorithm}
\DecMargin{1em}

Lines $1-3$ initialize the variables and insert all samples in the priority queue $Q$. The main loop, implemented in Lines $4-11$, is responsible for the conquering process. It first removes a sample $\bm{q}$ from $Q$ with maximum path value in Line $5$: if $\bm{q}$ has not been conquered by any other sample (Line $6$), it is then considered a prototype. Therefore, its connectivity value is reset to $\rho(\bm{q})$ (Line $7$) and a new distinct cluster label is assigned to it for optimum-path propagation. The inner loop in Lines $8-12$ evaluates samples in the neighborhood of $\bm{q}$ to which $\bm{q}$ can offer a better connectivity value. In this case (Line $10$), then the predecessor, label and cost maps of $\bm{u}$ are updated accordingly (Lines $11-12$).

Considering the implementation provided by Algorithm~\ref{a.unsupopf}, there is a need to first find $k^\ast\in[1,k_{max}]$ that represents the size of the neighborhood used to compute the density of each node (Equation~\ref{e.density}). Rocha et al.~\cite{RochaIJIST:09} proposed to find $k^\ast$ that minimizes the graph cut over the training set. Using any standard data structure, such an operation takes $\theta(\left|{\cal V}\right|^2)$ steps, and the whole process takes $\theta(k_{max}\left|{\cal V}\right|^2)$ operations. Since $k_{max}\in O(\left|{\cal V}\right|)$ is usually enough in most situations, the final complexity to find $k^\ast$ is given by $\theta(\left|{\cal V}\right|^2)$.

Lines $1-3$ take $\theta(\left|{\cal V}\right|)$ operations, while the loop in Lines $4-12$ removes each sample only once. Therefore, it runs over $\theta(\left|{\cal V}\right|)$ iterations. Implementing the priority queue $Q$ as a binary heap, we can execute Line $5$ in $O(\log\left|{\cal V}\right|)$ steps. Besides, Lines $8-12$ take $O(k^\ast)$ operations, thus ending up in $O(\left|{\cal V}\right|(\log\left|{\cal V}\right|+k^\ast) = O(\left|{\cal V}\right|\log\left|{\cal V}\right|)+O(k^\ast\left|{\cal V}\right|)$ operations concerning the main loop. In practice, this complexity nails down to $O(\left|{\cal V}\right|\log\left|{\cal V}\right|)$. The final complexity concerning unsupervised OPF is given by $\theta(\left|{\cal V}\right|^2)+O(\left|{\cal V}\right|\log\left|{\cal V}\right|)$, which is in practice $\theta(\left|{\cal V}\right|^2)$.

\section{Proposed Approaches}
\label{s.approaches}

This section describes the proposed framework based on the Optimum-Path Forest to tackle imbalanced datasets using oversampling and undersampling strategies.

\subsection{Undersampling with Optimum-Path Forest}
\label{ss.OPF-US}

The OPF-US employs a straightforward concept of ranking samples according to their importance during training for further pruning the less relevant ones. The whole process can be defined into two main steps: (i) training standard OPF using Algorithm~\ref{a.opf_sup}, and then (ii) assigning a score to each training sample during a validation step.  

Assigning scores is straightforward using the OPF algorithm, as demonstrated by Fernandes and Papa~\cite{FernandesPAA:19}. During validation, we computed how many times a training sample labeled a validation node correctly. On the other hand, in the case of misclassification, that training sample will have its score penalized. Basically, training samples are assigned zero scores, which are then incremented by one unit in case of correct classification, or decreased by one unit otherwise. In the end, each training sample will possess a score that somehow encodes its importance in the conquering process. Finally, samples are ranked in ascending order considering their scores, and the less-ranked samples are pruned from the training set until it is fully balanced, i.e., the number of samples from the two classes are equal. Notice only samples from the majority class are pruned. Additionally, such an approach considers binary datasets only, but it can be readily extended to multi-class scenarios. 

\IncMargin{1em}
\begin{algorithm}[!ht]
\SetKwInput{KwInput}{Input}
\SetKwInput{KwOutput}{Output}
\SetKwInput{KwAuxiliary}{Auxiliary}

\caption{OPF-US Algorithm}
\label{a.OPF-US}
\Indm
\KwInput{A training set ${\cal V}$, a validation set ${\cal T}$, map of training set labels $\lambda$, map of validation set labels $\omega$, number of samples to be removed $n_r$, set of prototypes ${\cal P}^\ast \subseteq {\cal V}$, and distance function $d$.}
\KwOutput{Undersampled training set ${\cal V}^{\star}$.}
\KwAuxiliary{Score map $S(\bm{q})=0, \forall \bm{q}\in{\cal V}$, predecessor $O^\prime$, cost $C^\prime$ and label $L^\prime$ maps concerning the training set, label map $L$ regarding the validation set, as well as the lower-scored majority samples from training set $\hat{\cal V}$.}
\Indp\Indpp

\BlankLine
$[O^{\prime},C^{\prime},L^{\prime}]\leftarrow$ Algorithm~\ref{a.opf_sup}(${\cal V}, {\cal P}^\ast, \lambda,d$)\;

\For{each $\bm{t}\in{\cal T}$}{
	Let $\bm{s}^\ast\in{\cal V}$ be the sample that conquers $\bm{t}$ (Equation~\ref{e.conquering_function})\;
	$L(\bm{t})\leftarrow L^\prime(\bm{s}^\ast)$\;
	\If{$\omega(\bm{t})=L(\bm{t})$}{
		$S(\bm{s}^\ast)\leftarrow S(\bm{s}^\ast)+1$;}
	\lElse{$S(\bm{s}^\ast)\leftarrow S(\bm{s}^\ast)-1$}
}

$\hat{\cal V}\leftarrow$ $n_r$ lower-scored majority class samples from ${\cal V}$ according to $S$\;
${\cal V}^{\star}\leftarrow {\cal V}\backslash\hat{\cal V}$\;
\Return ${\cal V}^{\star}$
\end{algorithm}
\DecMargin{1em}

Algorithm~\ref{a.OPF-US} implements the proposed method for undersampling purposes. Line $1$ performs the training step of standard OPF, while the loop in Lines $2-7$ provides a given training sample $\bm{s}^\ast$ a score. Line $3$ finds the training sample that conquers $\bm{t}$, and Line $4$ assigns the label of $\bm{s}^\ast$ to it. Line $5$ checks whether sample $\bm{t}$ has been classified correctly (score is increased in Line $6$) or not (score is decreased in Line $7$). Later, in Line $8$, $\hat{\cal V}$ receives the $n_r$ lower-scored samples of the majority class from ${\cal V}$, in which $n_r$ stands for the difference between the number of samples from the majority and minority classes. Finally, in Line $9$, such samples are removed from the original training dataset. 

\begin{figure}[!htb]
  \centerline{
    \begin{tabular}{c}
		\includegraphics[width=.6\textwidth]{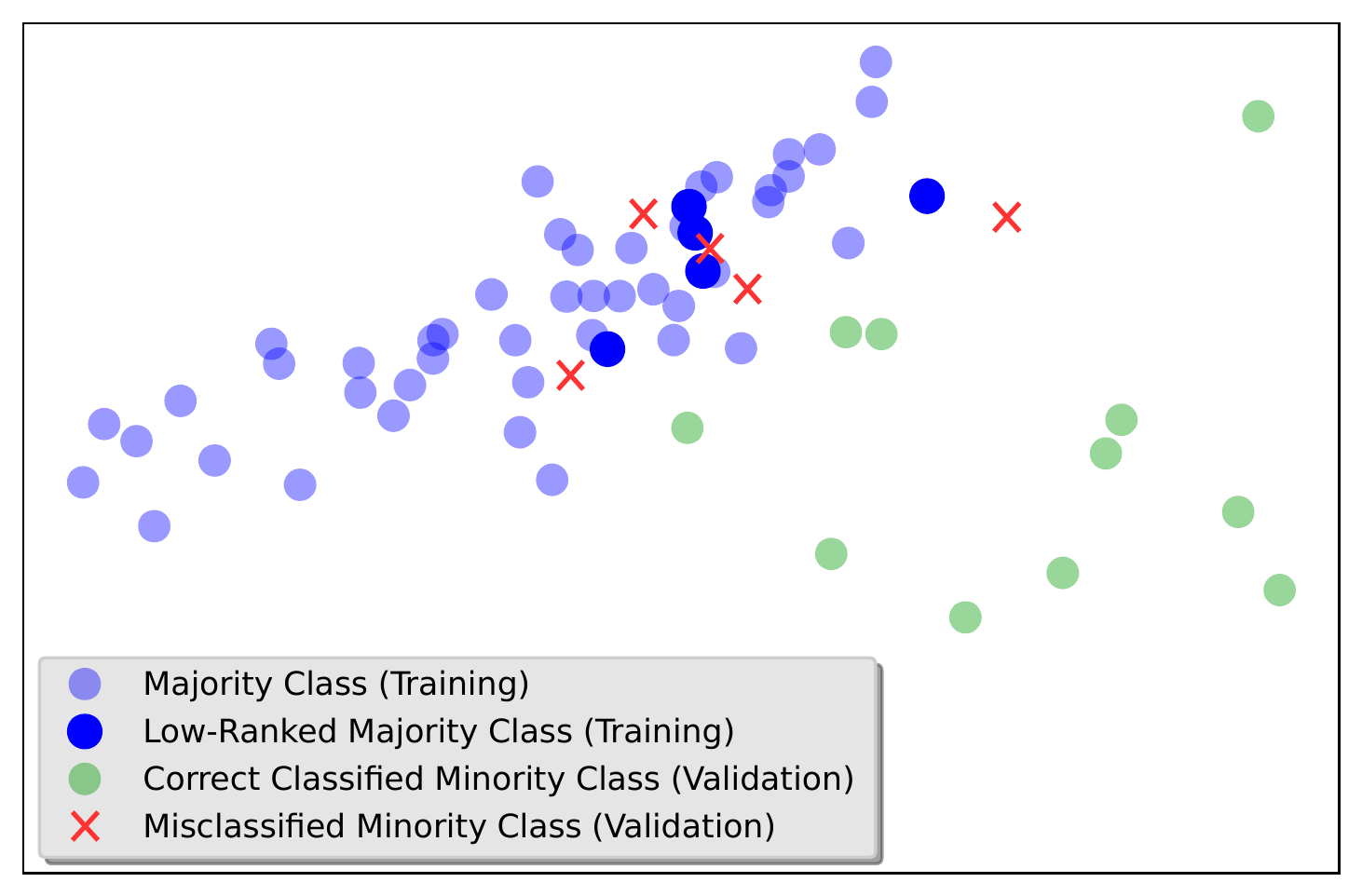} 
    \end{tabular}}
    \caption{Illustration of the OPF-US process: translucent blue dots denote the training set standard majority class samples, while solid blue dots stand for the ones that misclassfied some minority class samples from the validation set (illustrated as `X' symbols in red). Light green dots represent minority class samples from the validation set classified correctly. Notice that solid blue samples are likely to be pruned in the proposed undersampling procedure.}
  \label{f.opfus}
\end{figure}

Figure~\ref{f.opfus} glances the rationale behind the OPF-US working mechanism, where the blue dots stand for the majority class samples. The main idea is to remove training samples that do not contribute or add little to the classification of unseen samples (represented here as the green dots). Solid blue dots denote low-ranked samples, which are likely to be located in regions that figure some degree of overlap between classes. Therefore, by pruning such samples, we expect the number of misclassifications to drop considerably.

\subsubsection{OPF-US Computational Complexity}
\label{sss.OPF-USComplexity}

OPF-US complexity is, in large part, governed by the OPF classifier complexity. According to Section~\ref{ss.supOPF}, Line $1$ takes $\theta(\left|{\cal V}\right|)$ steps. Further, Lines $2-7$ implement the main loop, which requires $\theta(\left|{\cal T}\right|)$ runs. Besides, Line $3$ takes $O(\left|{\cal V}\right|)$ steps, while the remaining lines (i.e., $4-7$) figure a constant complexity, i.e., $\theta(1)$. Therefore, the whole loop requires $O(\left|{\cal V}\right|.\left|{\cal T}\right|)$ computations.

Line $8$ extracts the majority samples from ${\cal V}$ according to their scores, thus generating a smaller set $\hat{\cal V}$. Samples are sorted using any traditional algorithm with complexity $\theta(\left|{\cal V}\right|\log\left|{\cal V}\right|)$, for further removing the $n_r$ lower-scored samples to $\hat{\cal V}$. This last step takes $\theta(n_r)$. In practice, Line $8$ takes $\theta(\left|{\cal V}\right|\log\left|{\cal V}\right|)$ steps. Finally, in Line $9$, ${\cal V}^{\star}$ receives ${\cal V}\backslash\hat{\cal V}$, which is performed in $\theta(\left|{\cal V}\right|-\hat{\cal V}|)$ steps since both sets are sorted already. The final complexity of the OPF-US algorithm is then given by $\theta(\left|{\cal V}\right|)^2+\theta(\left|{\cal V}\right|\log\left|{\cal V}\right|)$, i.e., $\theta(\left|{\cal V}\right|)^2$.

\subsubsection{OPF-US Variations}
\label{sss.oupfVariations} 

Despite the naive OPF-US, we also propose three variations for undersampling purposes using the Optimum-Path Forest: (i) OPF-US1; (ii) OPF-US2; and the (iii) OPF-US3. These techniques are described as follows:

\begin{enumerate}
	\item \textbf{OPF-US1}: remove samples from the majority class with negative scores only;
	\item \textbf{OPF-US2}: remove samples from the majority class with scores lower or equal to zero; and
	\item \textbf{OPF-US3}: remove all samples with negative scores.
\end{enumerate}

\subsection{Oversampling with Optimum-Path Forest}
\label{ss.o2pf}

Additionally to the OPF-US, the OPF algorithm can also be modified to perform oversampling, hereinafter called Oversampling Optimum-Path Forest ($\text{O}^2$PF). In this case, one is concerned with generating new training examples for the minority class while enforcing their diversity and plausibility, i.e., new samples should share relevant characteristics with the real ones. To fulfill that purpose, $\text{O}^2$PF employs the unsupervised OPF to group training samples first and then estimate a Gaussian distribution for each cluster\footnote{Clustering by OPF considers the groups can be modeled by Gaussian distributions, but it is not limited to that.}. With the parameters of the model in hand, we can generate new samples accordingly.

Let ${\cal B}= \{b_1,b_2,\dots,b_c\}$ be the set of clusters computed by unsupervised OPF, as described in Section~\ref{ss.unsupOPF}. Also, Let $n_s$ be the number of samples to be generated by the model, which are distributed proportionally to the cluster's size (i.e., the number of samples). Therefore, the more dense the cluster, the higher the number of synthetic samples generated within it. As such, some low-density clusters may not be able to produce new samples. Such steps aim at fostering diversity among to-be-created examples while keeping their distribution. Moreover, a new sample $\bm{z}$ can be generated as follows:

\begin{equation}
	\bm{z} \sim \mathcal{N}(\bm{\mu}_i, \bm{\Sigma}_i),
	\label{e.sampling}
\end{equation}
where $\bm{\mu}_i$ and $\bm{\Sigma}_i$ denote the parameters of cluster $b_i\in{\cal B}$.

To summarize, the proposed $\text{O}^2$PF algorithm consists of three phases:
\begin{enumerate}
\item To identify each cluster within the class to be oversampled.
\item For each cluster $b_i\in{\cal B}$, we compute its mean vector $\bm{\mu}_j$ and covariance matrix $\bm{\Sigma}_j$.
\item To generate new samples for each cluster based on the previous parameters.
\end{enumerate}
It is worth mentioning that the last two steps can be parameterized, yielding variants of the proposed method, as discussed in Section~\ref{sss.o2pfVariations}. Algorithm~\ref{a.o2pf} implements $\text{O}^2$PF. 

\IncMargin{1em}
\begin{algorithm}
\SetKwInput{KwInput}{Input}
\SetKwInput{KwOutput}{Output}
\SetKwInput{KwAuxiliary}{Auxiliary}

\caption{$\text{O}^2$PF Algorithm}
\label{a.o2pf}
\Indm
\KwInput{Graph ${\cal G}_s = ({\cal L}, {\cal A}_k)$ containing the minority class training samples such that ${\cal L}\subset{\cal V}$, and the number $n_s$ of synthetic samples to be generated.}
\KwOutput{Set of minority class synthetic samples ${\cal L}^\prime$.}
\KwAuxiliary{Predecessor map $O$, path-cost map $C$, and cluster label map $L$, and variables $tmp$ and $ctr$.}
\Indp\Indpp

\BlankLine
$[O,C,L]\leftarrow$Algorithm 2$({\cal G}_s)$\;
${\cal L}^\prime\leftarrow \emptyset$; $ctr\leftarrow0$; ${\cal B}\leftarrow$ load clusters\;
\For {each $i\in \{1,2,\ldots,\left|{\cal B}\right|\}$ and $ctr<n_s$}{
	Compute the mean $\bm{\mu}_i$ and covariance matrix $\bm{\Sigma}_i$\;
	$tmp\leftarrow \left\lfloor\frac{\left|b_i\right|}{\left|{\cal L}\right|}\times n_s\right\rfloor$\;
	\For {$1$ to $tmp$}{
		Generate sample $\bm{z}\sim \mathcal{N}(\bm{\mu}_i, \bm{\Sigma}_i)$\;
		${\cal L}^\prime\leftarrow {\cal L}^\prime\cup\{\bm{z}\}$; $ctr\leftarrow ctr+1$\;
	}
}
\Return ${\cal L}^\prime$
\end{algorithm}
\DecMargin{1em}

Line $1$ performs data clustering using unsupervised OPF, and Line $2$ initializes the set of synthetic samples and the set of clusters. The main loop (Lines $3-8$) is responsible for computing the cluster parameters (i.e., mean and covariance matrix), the number of samples to be generated according to the cluster's size (Line $5$), and sampling from the distributions (Lines $6-8$). Finally, Line $9$ outputs the set of generated samples ${\cal L}^\prime$.

\subsubsection{$\text{O}^2$PF Computational Complexity}
\label{sss.o2pfComplexity}

Considering the implementation provided by Algorithm~\ref{a.o2pf}, Line $1$ takes $\theta(\left|{\cal L}\right|^2)$ operations according to Section~\ref{ss.unsupOPF}, while Line $2$ takes $\theta(\left|{\cal L}\right|)$ operations using a linked list. The main loop in Lines $3-9$ runs over $O(\left|{\cal B}\right|)$ iterations, and the inner loop in Lines $6-8$ takes $n_s$ steps only once. Similarly, Line $4$ also takes $\theta(\left|{\cal L}\right|)$ steps only once. Therefore, the complexity of Lines $3-8$ is given by $O(\theta(\left|{\cal L}\right|+n_s)$. Last but not least, the final complexity of $\text{O}^2$PF is given by $\theta(\left|{\cal L}\right|^2) + O(\theta(\left|{\cal L}\right|+n_s)$, i.e., $\theta(\left|{\cal L}\right|^2)$.

\subsubsection{$\text{O}^2$PF Variations}
\label{sss.o2pfVariations}

Unsupervised learning by OPF aims to cluster the dataset in such a way that groups of samples that share some degree of similarity are connected in the feature space. Standard $\text{O}^2$PF estimates a Gaussian distribution for each cluster computing the mean value and the covariance of the samples within that cluster. In this paper, we also explored different methodologies to model the data distribution for each cluster, as well as the approach to sampling new data. As such, this paper also proposes four $\text{O}^2$PF variants, described as follows:

\begin{enumerate}
	\item \textbf{$\text{O}^2$PF Radius Interpolation ($\text{O}^2$PF$_{RI}$)}: it replaces the cluster mean by its geometric median $\bm{g}$, thus ideally making the model more robust to outliers. Further, the sampling process is replaced by a radius interpolation, i.e., each new instance is generated by randomly selecting the $r$-th sample $\bm{x}_r\in{\cal V}$ in the cluster and interpolating it as follows: ${\bm{z}^\prime = \beta \bm{x}_r} + (1 - \beta)\bm{g}$, such that ${\beta \sim \mathcal{U}(0, \left[1 + d(\bm{g}, \bm{x}_{r})\right]^{-1})}$. Notice $\bm{g}$ is updated after the generation of each new sample\footnote{In this paper, $d(\cdot,\cdot)$ stands for the Euclidean distance, but any other distance/similarity function can be used.}.
	\item \textbf{$\text{O}^2$PF Mean Interpolation ($\text{O}^2$PF$_{MI}$)}: after generating the new sample $\bm{z}$, it is further interpolated with its nearest neighbor $\bm{p}$ within the cluster as follows: $\bm{z}^\prime = (1 - \alpha)\bm{p} + \alpha \bm{z}$, in which $\alpha \sim \mathcal{U}(0, 1)$.
	\item \textbf{$\text{O}^2$PF Prototype ($\text{O}^2$PF$_{P}$)}: instead of using the cluster's mean as a parameter of the Gaussian distribution, as employed by standard $\text{O}^2$PF, we use the prototype sample of that cluster instead. The sampling process remains the same one used by $\text{O}^2$PF.
	\item \textbf{$\text{O}^2$PF Weight Interpolation ($\text{O}^2$PF$_{WI}$)}: it employs a strategy that weights each sample according to its density (Equation~\ref{e.density}). Therefore, the cluster's mean stands for the weighted average of its samples. The sampling process remains the same one used by $\text{O}^2$PF$_{MI}$.
\end{enumerate}

\subsection{Hybrid approaches}
\label{ss.hybrid}

Despite the approaches introduced in this work, we also propose three hybrid models that combine data undersampling and oversampling. The rationale behind these models is to first remove less important samples (i.e., instances that may lead to misclassification) for further generating new data that is likely to be more reliable and adherent to the data distribution.

\begin{enumerate}
	\item \textbf{OPF-US1-$\text{O}^2$PF}: it removes samples from majority class with negative scores using OPF-US1 for further oversampling using $\text{O}^2$PF.
	\item \textbf{OPF-US2-$\text{O}^2$PF}: it removes samples from majority class with negative scores using OPF-US2 for further oversampling using $\text{O}^2$PF.
	\item \textbf{OPF-US3-$\text{O}^2$PF}: it removes all samples with negative scores using OPF-US3 for further oversampling using $\text{O}^2$PF.
\end{enumerate}

\section{Methodology}
\label{s.methodology}

This section presents the datasets and the experimental setup adopted in the paper.

\subsection{Datasets}
\label{ss.datasets}

The experiments are evaluated over three groups of applications that usually suffer from imbalanced datasets: (i) dissolved gas analysis (comprises faults on power transformers), (ii) a collection of medical-related datasets obtained from UCI~\cite{Dua:2019}, and (iii) a compilation of general-purpose datasets, also obtained from UCI.

\subsubsection{Dissolved Gas Analysis Datasets}
\label{sss.electrical}

The Dissolved Gas Analysis (DGA) in insulating oil is a widely used method for incipient fault interpretation in large oil-filled power transformers. For fault diagnosis, there are $7$ main attributes to be observed, which are types of gases produced by oil and cellulose decompositions from thermal and electrical stresses that occur with equipment. The data used to create this dataset was obtained from IEC TC10~\cite{duval2001interpretation}, a Brazilian private power distribution company~\cite{lupi2012comparaccao,ghoneim2012dissolved,soni2016approach}, as well as from scientific articles concerning DGA analysis~\cite{equbal2018transformer}. The data is divided into three main classes: normal (i.e., no-fault) samples ($1,012$ instances), thermal fault ($57$ instances), and electrical fault ($74$ instances). Finally, this data is combined into six binary fault subsets to evaluate the proposed method, described as follows:

\begin{itemize}
  \item \textbf{1069\_5gt} and \textbf{1069\_7gt:} both subsets are composed of $1,069$ samples, where $1,012$ ($94.67\%$) denote normal samples, while $57$ ($5.33\%$) represent thermal faults\footnote{\label{fn.gas}The 1069\_5gt, 1086\_5ge, and 1143\_5gte datasets comprise five types of gases, i.e., $H_2$, $CH_4$, $C_2H_2$, $C_2H_4$, and $C_2H_6$. On the other hand, the 1069\_7gt, 1086\_7ge, and 1143\_7gte datasets are composed of the five gases mentioned before plus $CO$ and $CO_2$.}.
	\item \textbf{1086\_5ge} and \textbf{1086\_7ge:} both subsets are composed of $1,086$ samples, where $1,012$ ($93.19\%$) denote normal samples, while $74$ ($6.81\%$) represent electrical faults\cref{fn.gas}.
	\item \textbf{1143\_5gte} and \textbf{1143\_7gte:} both subsets are composed of $1,143$ samples, where $1,012$ ($93.19\%$) denote normal samples, while $131$ ($11.46\%$) represent thermal and electrical faults\cref{fn.gas}.
\end{itemize}

\subsubsection{Medical-related Datasets}
\label{sss.medical}

We know that most medical datasets suffer from imbalance. In this paper, we compiled several public datasets from different applications, described as follows:

\begin{itemize}
	\item \textbf{Wisconsin Breast Cancer Database (WBCD) Prognostic}\footnote{Available at \url{https://archive.ics.uci.edu/ml/datasets/Breast+Cancer+Wisconsin+(Prognostic)}.}\textbf{:} it regards predicting a sample as recurrent or non-recurrent type of cancer based on $32$ features. There are $198$ samples, being $151$ ($76.3\%$) non-recurrent and $47$ ($23.7\%$) recurrent.
	\item \textbf{WBCD Diagnostic I}\footnote{Available at \url{http://archive.ics.uci.edu/ml/datasets/breast+cancer+wisconsin+(diagnostic)}.}\textbf{:} it concerns classifying a tumor as benignant or malignant based on $32$ features as well. There are $569$ instances, from which $357$ ($63.7\%$) are benign and $212$ ($37.3\%$) are malignant.
	\item \textbf{WBCD Diagnostic II}\footnote{Available at \url{http://archive.ics.uci.edu/ml/datasets/Breast+Cancer+Wisconsin+(Original)}.}\textbf{:} it corresponds to labeling each instance from a total of $699$ samples as benign or malignant tumor. Each sample comprises $9$ features and each class contains $458$ ($65.5\%$) and $241$ ($34.5\%$) samples, respectively.
	\item \textbf{Diabetic Retinopathy Debrecen (DRD)}\footnote{Available at \url{https://archive.ics.uci.edu/ml/datasets/Diabetic+Retinopathy+Debrecen+Data+Set}.}\textbf{:} it regards predicting whether an image contains signs of diabetic retinopathy or not based on $19$ variables. The dataset contains $1,151$ samples, from which $611$ ($53.1\%$) are positive and $540$ ($46.9\%$) are negative.
	\item \textbf{Cervical Cancer (CC)}\footnote{Available at \url{https://archive.ics.uci.edu/ml/datasets/Cervical+cancer+\%28Risk+Factors\%29}.}\textbf{:} it comprises $858$ samples from biopsies composed of $32$ features each. The dataset is significantly skewed, with $55$ $(6.4\%)$ positive and $803$ ($93.6\%$) negative samples.
	\item \textbf{Mammographic Mass (MM)}\footnote{Available at \url{https://archive.ics.uci.edu/ml/datasets/Mammographic+Mass}.}\textbf{:} it concerns predicting whether a mammographic mass is benign or malignant based on $6$ features. The dataset contains $516$ ($53.7\%$) benign and $445$ ($46.3\%$) malignant samples, comprising $961$ samples.
\end{itemize}

\subsubsection{General-Purpose Datasets}
\label{sss.general}

We also considered a set of datasets from different domains to evaluate the proposed approaches, as follows:

\begin{itemize}
  \item \textbf{Indian Liver (IL)}\footnote{Available at \url{https://archive.ics.uci.edu/ml/datasets/ILPD+(Indian+Liver+Patient+Dataset)}.}\textbf{:} it consists of predictions of liver diseases in $583$ individuals. The dataset is composed of $416$ patients and $167$ individuals that are affected by diseases. Each sample is composed of $10$ numerical attributes.
  \item \textbf{Secom}\footnote{\url{https://archive.ics.uci.edu/ml/datasets/SECOM}.}\textbf{:} it is omposed of $1,567$ samples from a semiconductor manufacturing process with $591$ features each, where $104$ samples denote fail in a manufacturing testing procedure, and the remaining $1,463$ passed in the test.
  \item \textbf{Seismic Bumps (SB)}\footnote{\url{https://archive.ics.uci.edu/ml/datasets/seismic-bumps}.}\textbf{:} it comprises $2,584$ samples with $19$ features from predictions of dangerous seismic activities in underground coal mining, where $2,414$ samples stand for negative predictions and $170$ for positive ones.
	\item \textbf{Spam}\footnote{\url{https://archive.ics.uci.edu/ml/datasets/Spambase}.}\textbf{:} it regards spam classification in e-mails. This dataset is slightly imbalanced, from which $2,788$ are non-spam samples ($60.6\%$) and $1,813$ samples represent spam ($39.4\%$), totaling $4,601$ samples and $57$ numerical attributes.
  \item \textbf{Vertebral Column (VC)}\footnote{\url{https://archive.ics.uci.edu/ml/datasets/Vertebral+Column}.}\textbf{:} it concerns the classification of orthopedic exams from column injuries comprising $100$ samples from regular ($32.3\%$) and $210$ from abnormal column ($67.7\%$), totaling $310$ samples with $6$ numerical attributes each.
  \item \textbf{Wilt}\footnote{\url{https://archive.ics.uci.edu/ml/datasets/Wilt}.}\textbf{:} it regards the prediction of diseased trees using numerical attributes acquired from a series of Quickbird images. The dataset contains $4,839$ samples with $5$ features each, where $4,578$ samples relate to non-diseased trees and only $261$ samples of diseased ones. 
\end{itemize}

\subsection{Experimental Setup}
\label{ss.experimental}

The paper employs three main approaches for dealing with imbalanced datasets through OPF-based methods, i.e., oversampling, undersampling, and hybrid methodologies considering undersampling followed by oversampling. For each group of datasets, we compare the standard $\text{O}^2$PF and its variants against seven state-of-the-art methods for dataset oversampling, i.e., SMOTE~\cite{ChawlaJAIR:02}, borderline SMOTE~\cite{han:05borderline}, ADASYN~\cite{He:2008adasyn}, AHC~\cite{Cohen2006AHC}, MWMOTE~\cite{Barua2014MWMOTE}, SOMO~\cite{douzas2017self}, and KMeansSMOTE~\cite{Douzas2018KmeansSmote}. Similarly, we also compare standard OPF-US and its three variants against five state-of-the-art baselines considered for data undersampling, i.e., CNN~\cite{hart1968condensed}, NearMiss-1~\cite{maniWLID2003}, NearMiss-2~\cite{maniWLID2003}, NearMiss-3~\cite{maniWLID2003} and Radial-Based Undersampling~\cite{koziarski2020radial}. Finally, the techniques that obtained the best results are compared against themselves and three other hybrid methods that combine OPF-US and $\text{O}^2$PF. The work adopted the well-known F1 score for comparison purposes since it considers a weighted average of the precision and recall, thus proper for evaluating imbalanced datasets. Besides, we used standard datasets as baselines, and the recognition rates are computed using the OPF classifier since it is a parameterless and determinist approach. However,  any other supervised technique could be used as well.

Regarding the data pre-processing step, datasets with missing values were filled with the average value of those respective features. Further, all datasets were normalized using the standard scaler, which subtracts the mean from each sample and scales it to unit variance. Moreover, all datasets were randomly split into training, validation, and testing sets, with proportions of $70\%$, $15\%$, and $15\%$, respectively. Such a configuration was chosen empirically.

Considering the $\text{O}^2$PF, the validation set is employed to fine-tune $k_{max}\in[5,10,20,30,40,50]$ parameter that maximizes the performance. A similar approach was employed to find the best number of neighbors $k\in[5,6,7,8,9,10]$ considering SMOTE, borderline SMOTE, ADASYN, MWMOTE, and \textit{k}-means SMOTE. Regarding the $k$-means SMOTE, the optimal number of clusters $c\in[1,2,3,4,5,6,7,8,9,10]$ was  found to maximize the performance of the model. Regarding the Radial-based undersampling method, we adjusted parameter $\gamma\in[0.001, 0.01, 0.1, 1.0, 10.0, 100.0]$ in a similar fashion to the one proposed by Koziarski~\cite{koziarski2020radial}. Besides, SOMO employs the best set of parameters defined by the authors, and AHC is parameterless. 

For each dataset, the whole process was repeated during $20$ runs to compute the mean recognition rates and standard deviations, and further perform statistical analysis using the Wilcoxon signed-rank test~\cite{Wilcoxon:45} with $5\%$ of significance. Finally, the code employed in this work was implemented in Python\footnote{Available at: \url{https://github.com/Leandropassosjr/OpfImb}} and based on the OPFython library~\cite{rosa2020opfython,rosa2021opfython}.

\section{Experiments}
\label{s.experiments}

This section discusses the experiments using the methodology presented earlier concerning the proposed oversampling and undersampling approaches. For the sake of explanation, the bolded values denote the most accurate F1 scores according to the Wilcoxon signed-rank test with $5\%$ of significance. Similarly, values underlined stand for the highest average F1 score overall  (absolute value) considering each dataset.

\subsection{Oversampling Results}
\label{ss.oversamplingResults}

Results shown in Table~\ref{t.results_OversampleF1_electrical} indicate the mean F1 score obtained after applying the oversampling approaches to the problem of fault detection in power transformers.  One can observe the proposed approaches performed similarly to the state-of-the-art baseline techniques. One can highlight $\text{{O}}^2$PF and $\text{{O}}^2$PF$_{RI}$, as well as two baselines, i.e., AHC and SOMO, which obtained the best results in five out-of-six datasets. Moreover, half of the best overall results were obtained by $\text{{O}}^2$PF-based approaches, i.e., over 1069\_5gt, 1143\_5gte, and 1143\_7gte datasets, while ADASYN, MWMOTE, and $k$-means SMOTE produced the less significative performances. However, one advantage of $\text{{O}}^2$PF and its variants concerns their number of parameters, as presented in Table~\ref{t.number_parameters_o2pf}. Although SMOTE and ADASYN figure only one parameter, they did not perform well in this context.

\begin{table}[htb]
\caption{Number of parameters for each oversampling technique.}
\begin{center}
\renewcommand{\arraystretch}{1.5}
\setlength{\tabcolsep}{6pt}
\resizebox{0.35\textwidth}{!}{
\begin{tabular}{c|c}
\hhline{-|-|}
\hhline{-|-|}
\hhline{-|-|}
{\cellcolor[HTML]{EFEFEF}{\textbf{Technique}}}  & {\cellcolor[HTML]{EFEFEF}{\textbf{Number of parameters}}}\\ \hline
\textbf{OPF-based ones} & $1$\\\hline\hline
\textbf{SMOTE~\cite{ChawlaJAIR:02}} & $1$\\\hline
\textbf{Borderline SMOTE~\cite{han:05borderline}} & $2$\\\hline
\textbf{ADASYN~\cite{He:2008adasyn}} & $1$\\\hline
\textbf{AHC~\cite{Cohen2006AHC}} & $0$\\\hline
\textbf{MWMOTE~\cite{Barua2014MWMOTE}} & $3$\\\hline
\textbf{SOMO~\cite{douzas2017self}} & $1$\\\hline
\textbf{$k$-means SMOTE~\cite{Douzas2018KmeansSmote}} & $2$ \\
\hhline{-|-|}
\hhline{-|-|}
\hhline{-|-|}
\end{tabular}}
\label{t.number_parameters_o2pf}
\end{center}
\end{table}


\begin{table*}[!htb]
\caption{Average F1 score values considering the oversampling tasks over DGA datasets.}
\begin{center}
\renewcommand{\arraystretch}{1.0}
\setlength{\tabcolsep}{6pt}
\resizebox{\textwidth}{!}{
\begin{tabular}{c|c|c|c|c|c|c}
\hhline{-|-|-|-|-|-|-|}
\hhline{-|-|-|-|-|-|-|}
\hhline{-|-|-|-|-|-|-|}
{\cellcolor[HTML]{EFEFEF}{\textbf{Technique / Dataset}}}  & {\cellcolor[HTML]{EFEFEF}{\textbf{1069\_5gt}}} & {\cellcolor[HTML]{EFEFEF}{\textbf{1069\_7gt}}} & {\cellcolor[HTML]{EFEFEF}{\textbf{1086\_5ge}}} & {\cellcolor[HTML]{EFEFEF}{\textbf{1086\_7ge}}} & {\cellcolor[HTML]{EFEFEF}{\textbf{1143\_5gte}}} & {\cellcolor[HTML]{EFEFEF}{\textbf{1143\_7gte}}}\\ \hline
\textbf{$\text{{O}}^2$PF} & $\bm{0.7282\pm0.1116}$ & $\bm{0.6249\pm0.1097}$ & $0.7328\pm0.1052$ & $\bm{0.5501\pm0.0950}$ & $\bm{0.7981\pm0.0526}$ & $\bm{0.6871\pm0.0822}$\\\hline
\textbf{$\text{O}^2$PF$_{MI}$} & $\bm{0.7198\pm0.1101}$ & $0.5982\pm0.1066$ & $0.6965\pm0.1068$ & $\bm{0.5478\pm0.1103}$ & $\bm{0.7866\pm0.0695}$ & $\bm{0.6812\pm0.0814}$\\\hline
\textbf{$\text{O}^2$PF$_{RI}$} & $\bm{0.7115\pm0.0896}$ & $\bm{0.6798\pm0.1005}$ & $0.7485\pm0.1089$ & $\bm{0.5453\pm0.1001}$ & $\underline{\bm{0.8095\pm0.0570}}$ & $\bm{0.6857\pm0.0875}$\\\hline
\textbf{$\text{O}^2$PF$_{P}$} & $\underline{\bm{0.7394\pm0.0961}}$ & $0.6091\pm0.1249$ & $0.6886\pm0.1366$ & $\bm{0.5629\pm0.1207}$ & $\bm{0.8029\pm0.0752}$ & $\bm{0.6918\pm0.0941}$\\\hline
\textbf{$\text{O}^2$PF$_{WI}$} & $0.7038\pm0.1090$ & $0.5925\pm0.0802$ & $0.6956\pm0.1190$ & $\bm{0.5596\pm0.1131}$ & $0.7843\pm0.0770$ & $\underline{\bm{0.6922\pm0.0885}}$\\\hline\hline
\textbf{SMOTE~\cite{ChawlaJAIR:02}} & $\bm{0.7053\pm0.0894}$ & $0.5352\pm0.0946$ & $0.6932\pm0.1402$ & $0.5257\pm0.1030$ & $0.7550\pm0.0774$ & $\bm{0.6597\pm0.0928}$\\\hline
\textbf{Borderline SMOTE~\cite{han:05borderline}} & $0.6892\pm0.0794$ & $0.5977\pm0.1151$ & $0.7119\pm0.1213$ & $\underline{\bm{0.5698\pm0.1066}}$ & $0.7469\pm0.0701$ & $\bm{0.6862\pm0.0812}$\\\hline
\textbf{ADASYN~\cite{He:2008adasyn}} & $0.6553\pm0.0845$ & $0.5065\pm0.1061$ & $0.6494\pm0.1297$ & $0.5013\pm0.0776$ & $0.7093\pm0.0855$ & $0.6448\pm0.0742$\\\hline
\textbf{AHC~\cite{Cohen2006AHC}} & $\bm{0.7110\pm0.0832}$ & $\bm{0.6690\pm0.1048}$ & $\bm{0.7500\pm0.1066}$ & $\bm{0.5645\pm0.0981}$ & $0.7915\pm0.0572$ & $\bm{0.6920\pm0.0786}$\\\hline
\textbf{MWMOTE~\cite{Barua2014MWMOTE}} & $\bm{0.7003\pm0.1162}$ & $0.5334\pm0.1073$ & $0.6779\pm0.1115$ & $0.5029\pm0.0788$ & $0.7477\pm0.0751$ & $0.6343\pm0.0739$\\\hline
\textbf{SOMO~\cite{douzas2017self}} & $\bm{0.7087\pm0.0915}$ & $\underline{\bm{0.6830\pm0.1035}}$ & $\underline{\bm{0.7659\pm0.0992}}$ & $\bm{0.5369\pm0.1063}$ & $0.7996\pm0.0590$ & $\bm{0.6714\pm0.0908}$\\\hline
\textbf{$k$-means SMOTE~\cite{Douzas2018KmeansSmote}} & $0.6824\pm0.1122$ & $0.5466\pm0.1045$ & $0.6941\pm0.1264$ & $0.5096\pm0.0888$ & $0.7338\pm0.0762$ & $\bm{0.6693\pm0.0832}$\\
\hhline{-|-|-|-|-|-|-|}
\hhline{-|-|-|-|-|-|-|}
\hhline{-|-|-|-|-|-|-|}
\end{tabular}}
\label{t.results_OversampleF1_electrical}
\end{center}
\end{table*}

Table~\ref{t.results_OversampleF1_medical} presents the results over medical datasets, in which $\text{{O}}^2$PF-based approaches obtained the best overall results in four-out-of-six datasets. Statistically speaking, the proposed approaches obtained the best results in all datasets. On the other hand, although MWMOTE performed better than before (i.e., Table~\ref{t.results_OversampleF1_electrical}), it comprises more parameters than any other technique.

\begin{table*}[!htb]
\caption{Average F1 score values considering the oversampling tasks over medical datasets.}
\begin{center}
\renewcommand{\arraystretch}{1.0}
\setlength{\tabcolsep}{6pt}
\resizebox{\textwidth}{!}{
\begin{tabular}{c|c|c|c|c|c|c}
\hhline{-|-|-|-|-|-|-|}
\hhline{-|-|-|-|-|-|-|}
\hhline{-|-|-|-|-|-|-|}
{\cellcolor[HTML]{EFEFEF}{\textbf{Technique / Dataset}}}  & {\cellcolor[HTML]{EFEFEF}{\textbf{Diagnostic}}} & {\cellcolor[HTML]{EFEFEF}{\textbf{Diagnostic II}}} & {\cellcolor[HTML]{EFEFEF}{\textbf{Prognostic}}} & {\cellcolor[HTML]{EFEFEF}{\textbf{DDR}}} & {\cellcolor[HTML]{EFEFEF}{\textbf{CC}}} & {\cellcolor[HTML]{EFEFEF}{\textbf{MM}}}\\ \hline
\textbf{$\text{{O}}^2$PF} & $\bm{0.9290\pm0.0227}$ & $\underline{\bm{0.9240\pm0.0290}}$ & $0.4034\pm0.1282$ & $\underline{\bm{0.6125\pm0.0354}}$ & $\bm{0.5666\pm0.1437}$ & $\bm{0.6497\pm0.0524}$\\\hline
\textbf{$\text{O}^2$PF$_{MI}$} & $\bm{0.9311\pm0.0288}$ & $0.9129\pm0.0329$ & $0.4096\pm0.1043$ & $0.5813\pm0.0476$ & $\bm{0.5667\pm0.1308}$ & $0.6382\pm0.0656$\\\hline
\textbf{$\text{O}^2$PF$_{RI}$} & $\bm{0.9321\pm0.0238}$ & $0.9143\pm0.0357$ & $\underline{\bm{0.4589\pm0.1064}}$ & $0.5804\pm0.0489$ & $0.5178\pm0.1362$ & $\bm{0.6616\pm0.0596}$\\\hline
\textbf{$\text{O}^2$PF$_{P}$} & $\bm{0.9312\pm0.0238}$ & $0.9126\pm0.0343$ & $0.4242\pm0.1430$ & $0.5784\pm0.0485$ & $0.5255\pm0.1401$ & $0.6333\pm0.0744$\\\hline
\textbf{$\text{O}^2$PF$_{WI}$} & $\bm{0.9324\pm0.0248}$ & $0.9175\pm0.0303$ & $\bm{0.4110\pm0.1186}$ & $0.5804\pm0.0495$ & $\underline{\bm{0.5774\pm0.1263}}$ & $\bm{0.6313\pm0.0819}$\\\hline\hline
\textbf{SMOTE~\cite{ChawlaJAIR:02}} & $\bm{0.9330\pm0.0198}$ & $\bm{0.9170\pm0.0331}$ & $0.4112\pm0.1330$ & $0.5799\pm0.0465$ & $\bm{0.5673\pm0.0900}$ & $\bm{0.6591\pm0.0545}$\\\hline
\textbf{Borderline SMOTE~\cite{han:05borderline}} & $0.9231\pm0.0278$ & $0.9080\pm0.0353$ & $0.4175\pm0.1266$ & $0.5803\pm0.0460$ & $\bm{0.5414\pm0.1109}$ & $\underline{\bm{0.6710\pm0.0479}}$\\\hline
\textbf{ADASYN~\cite{He:2008adasyn}} & $\bm{0.9264\pm0.0285}$ & $\bm{0.9196\pm0.0315}$ & $\bm{0.4366\pm0.1413}$ & $0.5734\pm0.0469$ & $\bm{0.5675\pm0.0905}$ & $\bm{0.6631\pm0.0567}$\\\hline
\textbf{AHC~\cite{Cohen2006AHC}} & $\bm{0.9340\pm0.0229}$ & $\bm{0.9210\pm0.0352}$ & $0.4150\pm0.1286$ & $\bm{0.5990\pm0.0421}$ & $\bm{0.5240\pm0.1209}$ & $\bm{0.6435\pm0.0588}$\\\hline
\textbf{MWMOTE~\cite{Barua2014MWMOTE}} & $\bm{0.9305\pm0.0264}$ & $\bm{0.9223\pm0.0302}$ & $0.4211\pm0.1147$ & $0.5827\pm0.0462$ & $\bm{0.5528\pm0.1215}$ & $0.6349\pm0.0643$\\\hline
\textbf{SOMO~\cite{douzas2017self}} & $\underline{\bm{0.9359\pm0.0246}}$ & $\bm{0.9167\pm0.0318}$ & $0.4028\pm0.1328$ & $0.5804\pm0.0478$ & $\bm{0.5121\pm0.1306}$ & $0.6224\pm0.0627$\\\hline
\textbf{$k$-means SMOTE~\cite{Douzas2018KmeansSmote}} & $\bm{0.9345\pm0.0273}$ & $0.9142\pm0.0298$ & $\bm{0.4197\pm0.1092}$ & $0.5757\pm0.0468$ & $\bm{0.5575\pm0.0774}$ & $\bm{0.6505\pm0.0556}$\\
\hhline{-|-|-|-|-|-|-|}
\hhline{-|-|-|-|-|-|-|}
\hhline{-|-|-|-|-|-|-|}
\end{tabular}}
\label{t.results_OversampleF1_medical}
\end{center}
\end{table*}

$\text{O}^2$PF$_{MI}$ obtained promising results over the general-purpose datasets, once it obtained the most accurate overall results in half of the datasets, as well as the best results in five-out-of-six datasets considering the Wilcoxon analysis, together with three other $\text{O}^2$PF-based approaches and four baselines, as shown in Table~\ref{t.results_OversampleF1_general}. Despite the statistical similarity, the average values obtained by $\text{O}^2$PF-based approaches are, in general, slightly better than the baselines.

\begin{table*}[!htb]
\caption{Average F1 score values considering the oversampling tasks over general-purpose datasets. Symbol `-' denotes the implementation provided did not performed properly.}
\begin{center}
\renewcommand{\arraystretch}{1.0}
\setlength{\tabcolsep}{6pt}
\resizebox{\textwidth}{!}{
\begin{tabular}{c|c|c|c|c|c|c}
\hhline{-|-|-|-|-|-|-|}
\hhline{-|-|-|-|-|-|-|}
\hhline{-|-|-|-|-|-|-|}
{\cellcolor[HTML]{EFEFEF}{\textbf{Technique / Dataset}}}  & {\cellcolor[HTML]{EFEFEF}{\textbf{Indian Liver}}} & {\cellcolor[HTML]{EFEFEF}{\textbf{Secom}}} & {\cellcolor[HTML]{EFEFEF}{\textbf{Seismic Bumps}}} & {\cellcolor[HTML]{EFEFEF}{\textbf{Spam}}} & {\cellcolor[HTML]{EFEFEF}{\textbf{Vertebral Column}}} & {\cellcolor[HTML]{EFEFEF}{\textbf{Wilt}}}\\ \hline
\textbf{$\text{{O}}^2$PF} & $\bm{0.4850\pm0.0566}$ & $\bm{0.1536\pm0.0357}$ & $\bm{0.1970\pm0.0877}$ & $\bm{0.8712\pm0.0144}$ & $\bm{0.6651\pm0.1290}$ & $0.5189\pm0.0507$\\\hline
\textbf{$\text{O}^2$PF$_{MI}$} & $\underline{\bm{0.4972\pm0.0615}}$ & $\bm{0.1572\pm0.0366}$ & $\underline{\bm{0.1987\pm0.0808}}$ & $\underline{\bm{0.8740\pm0.0132}}$ & $\bm{0.6524\pm0.1333}$ & $0.6162\pm0.0466$\\\hline
\textbf{$\text{O}^2$PF$_{RI}$} & $0.4516\pm0.0625$ & $\bm{0.1383\pm0.0340}$ & $0.1578\pm0.0784$ & $\bm{0.8683\pm0.0149}$ & $\bm{0.6518\pm0.1388}$ & $\bm{0.6599\pm0.0627}$\\\hline
\textbf{$\text{O}^2$PF$_{P}$} & $\bm{0.4791\pm0.0731}$ & $\underline{\bm{0.1664\pm0.0696}}$ & $\bm{0.1913\pm0.0779}$ & $\bm{0.8721\pm0.0128}$ & $\bm{0.6497\pm0.1400}$ & $0.5661\pm0.0891$\\\hline
\textbf{$\text{O}^2$PF$_{WI}$} & $\bm{0.4925\pm0.0489}$ & $\bm{0.1557\pm0.0375}$ & $\bm{0.1870\pm0.0695}$ & $\bm{0.8718\pm0.0177}$ & $\bm{0.6612\pm0.1323}$ & $0.6256\pm0.0460$\\\hline\hline
\textbf{SMOTE~\cite{ChawlaJAIR:02}} & $\bm{0.4699\pm0.0652}$ & $\bm{0.1577\pm0.0358}$ & $0.1581\pm0.0588$ & $\bm{0.8706\pm0.0155}$ & $\bm{0.6703\pm0.1276}$ & $\underline{\bm{0.6677\pm0.0498}}$\\\hline
\textbf{Borderline SMOTE~\cite{han:05borderline}} & $\bm{0.4769\pm0.0551}$ & $\bm{0.1542\pm0.0313}$ & $\bm{0.1751\pm0.0676}$ & $0.8664\pm0.0156$ & $\bm{0.6670\pm0.1369}$ & $\bm{0.6616\pm0.0471}$\\\hline
\textbf{ADASYN~\cite{He:2008adasyn}} & $\bm{0.4723\pm0.0738}$ & $\bm{0.1531\pm0.0390}$ & $0.1656\pm0.0674$ & $0.8664\pm0.0114$ & $\bm{0.6702\pm0.1386}$ & $\bm{0.6659\pm0.0555}$\\\hline
\textbf{AHC~\cite{Cohen2006AHC}} & $0.4645\pm0.0666$ & $\bm{0.1415\pm0.0294}$ & $0.1580\pm0.0786$ & $\bm{0.8710\pm0.0158}$ & $\bm{0.6680\pm0.1407}$ & $\bm{0.6575\pm0.0537}$\\\hline
\textbf{MWMOTE~\cite{Barua2014MWMOTE}} & $\bm{0.4700\pm0.0739}$ & $\bm{0.1501\pm0.0381}$ & $0.1754\pm0.0650$ & $\bm{0.8728\pm0.0183}$ & $\underline{\bm{0.6751\pm0.1217}}$ & $\bm{0.6593\pm0.0541}$\\\hline
\textbf{SOMO~\cite{douzas2017self}} & $0.4297\pm0.1224$ & - & - & $\bm{0.8729\pm0.0162}$ & $\bm{0.6623\pm0.1376}$ & - \\\hline
\textbf{$k$-means SMOTE~\cite{Douzas2018KmeansSmote}} & $\bm{0.4841\pm0.0601}$ & $\bm{0.1583\pm0.0363}$ & $0.1653\pm0.0603$ & $\bm{0.8704\pm0.0177}$ & $\bm{0.6625\pm0.1250}$ & $\bm{0.6607\pm0.0496}$\\
\hhline{-|-|-|-|-|-|-|}
\hhline{-|-|-|-|-|-|-|}
\hhline{-|-|-|-|-|-|-|}
\end{tabular}}
\label{t.results_OversampleF1_general}
\end{center}
\end{table*}

\subsection{Undersampling Results}
\label{ss.undersamplingResults}

Table~\ref{t.results_UndersampleF1_electrical} shows the results concerning the undersampling strategies over DGA datasets. The performance of OPF-US3 is paramount since it obtained the best overall results in all datasets. Additionally, one can notice that none of the baselines were capable of obtaining at least statistically similar results. Table~\ref{t.number_parameters_opf_us} presents the number of parameters of each technique, which is one of the main advantages regarding the supervised OPF classifier with complete graph.

\begin{table}[htb!]
\caption{Number of parameters for each undersampling technique.}
\begin{center}
\renewcommand{\arraystretch}{1.5}
\setlength{\tabcolsep}{6pt}
\resizebox{0.35\textwidth}{!}{
\begin{tabular}{c|c}
\hhline{-|-|}
\hhline{-|-|}
\hhline{-|-|}
{\cellcolor[HTML]{EFEFEF}{\textbf{Technique}}}  & {\cellcolor[HTML]{EFEFEF}{\textbf{Number of parameters}}}\\ \hline
\textbf{OPF-based ones} & $0$\\\hline\hline
\textbf{CNN~\cite{hart1968condensed}} & $1$\\\hline
\textbf{RBU~\cite{koziarski2020radial}} & $1$ \\\hline
\textbf{NearMiss-1~\cite{maniWLID2003}} & $1$ \\\hline
\textbf{NearMiss-2~\cite{maniWLID2003}} & $1$ \\\hline
\textbf{NearMiss-3~\cite{maniWLID2003}} & $1$ \\
\hhline{-|-|}
\hhline{-|-|}
\hhline{-|-|}
\end{tabular}}
\label{t.number_parameters_opf_us}
\end{center}
\end{table}


\begin{table*}[!htb]
\caption{Average F1 score values considering the undersampling tasks over DGA datasets.}
\begin{center}
\renewcommand{\arraystretch}{1.0}
\setlength{\tabcolsep}{6pt}
\resizebox{\textwidth}{!}{
\begin{tabular}{c|c|c|c|c|c|c}
\hhline{-|-|-|-|-|-|-|}
\hhline{-|-|-|-|-|-|-|}
\hhline{-|-|-|-|-|-|-|}
{\cellcolor[HTML]{EFEFEF}{\textbf{Technique / Dataset}}}  & {\cellcolor[HTML]{EFEFEF}{\textbf{1069\_5gt}}} & {\cellcolor[HTML]{EFEFEF}{\textbf{1069\_7gt}}} & {\cellcolor[HTML]{EFEFEF}{\textbf{1086\_5ge}}} & {\cellcolor[HTML]{EFEFEF}{\textbf{1086\_7ge}}} & {\cellcolor[HTML]{EFEFEF}{\textbf{1143\_5gte}}} & {\cellcolor[HTML]{EFEFEF}{\textbf{1143\_7gte}}}\\ \hline
\textbf{OPF-US} & $0.5250\pm0.0776$ & $0.3593\pm0.0900$ & $0.4846\pm0.0976$ & $0.3023\pm0.0825$ & $0.6645\pm0.0720$ & $0.5184\pm0.0951$\\\hline
\textbf{OPF-US1} & $\bm{0.7053\pm0.1063}$ & $\bm{0.6760\pm0.1076}$ & $0.7666\pm0.0964$ & $0.5357\pm0.1024$ & $0.7798\pm0.0750$ & $0.6748\pm0.0955$\\\hline
\textbf{OPF-US2} & $\bm{0.7224\pm0.1057}$ & $\bm{0.6488\pm0.1074}$ & $0.7663\pm0.0920$ & $0.5228\pm0.0951$ & $\bm{0.7802\pm0.0727}$ & $0.6450\pm0.0975$\\\hline
\textbf{OPF-US3} & $\underline{\bm{0.7274\pm0.0995}}$ & $\underline{\bm{0.6853\pm0.1131}}$ & $\underline{\bm{0.8071\pm0.0883}}$ & $\underline{\bm{0.5970\pm0.1073}}$ & $\underline{\bm{0.7969\pm0.0728}}$ & $\underline{\bm{0.6960\pm0.0964}}$\\\hline\hline
\textbf{CNN~\cite{hart1968condensed}} & $0.4214\pm0.2313$ & $0.2657\pm0.0894$ & $0.4653\pm0.1714$ & $0.3342\pm0.1576$ & $0.5003\pm0.1559$ & $0.4512\pm0.1123$\\\hline
\textbf{RBU~\cite{koziarski2020radial}} & $0.3699\pm0.2335$ & $0.2221\pm0.1515$ & $0.4112\pm0.2359$ & $0.1847\pm0.0868$ & $0.3619\pm0.2264$ & $0.3138\pm0.1268$\\\hline
\textbf{NearMiss-1~\cite{maniWLID2003}} & $0.3875\pm0.1482$ & $0.1467\pm0.0483$ & $0.2586\pm0.0556$ & $0.1384\pm0.0304$ & $0.4288\pm0.0905$ & $0.3007\pm0.0461$\\\hline
\textbf{NearMiss-2~\cite{maniWLID2003}} & $0.0862\pm0.0189$ & $0.1017\pm0.0366$ & $0.1306\pm0.0297$ & $0.1236\pm0.0336$ & $0.2063\pm0.0356$ & $0.2352\pm0.0665$\\\hline
\textbf{NearMiss-3~\cite{maniWLID2003}} & $0.2871\pm0.1712$ & $0.1334\pm0.0564$ & $0.1638\pm0.0387$ & $0.1247\pm0.0318$ & $0.4420\pm0.2282$ & $0.2727\pm0.0417$\\

\hhline{-|-|-|-|-|-|-|}
\hhline{-|-|-|-|-|-|-|}
\hhline{-|-|-|-|-|-|-|}
\end{tabular}}
\label{t.results_UndersampleF1_electrical}
\end{center}
\end{table*}

Results obtained over the medical datasets, provided in Table~\ref{t.results_UndersampleF1_medical}, describe a scenario pretty much similar to the one mentioned above, i.e., OPF-US-based approaches obtained the best results, with one standing out, i.e., OPF-US2. Moreover, as well as in the DGA datasets, none of the baselines were capable of obtaining statistically similar results.

\begin{table*}[!htb]
\caption{Average F1 score values considering the undersampling tasks over medical datasets.}
\begin{center}
\renewcommand{\arraystretch}{1.0}
\setlength{\tabcolsep}{6pt}
\resizebox{\textwidth}{!}{
\begin{tabular}{c|c|c|c|c|c|c}
\hhline{-|-|-|-|-|-|-|}
\hhline{-|-|-|-|-|-|-|}
\hhline{-|-|-|-|-|-|-|}
{\cellcolor[HTML]{EFEFEF}{\textbf{Technique / Dataset}}}  & {\cellcolor[HTML]{EFEFEF}{\textbf{Diagnostic}}} & {\cellcolor[HTML]{EFEFEF}{\textbf{Diagnostic II}}} & {\cellcolor[HTML]{EFEFEF}{\textbf{Prognostic}}} & {\cellcolor[HTML]{EFEFEF}{\textbf{DDR}}} & {\cellcolor[HTML]{EFEFEF}{\textbf{CC}}} & {\cellcolor[HTML]{EFEFEF}{\textbf{MM}}}\\ \hline
\textbf{OPF-US} & $\bm{0.9317\pm0.0236}$ & $\bm{0.9334\pm0.0228}$ & $\bm{0.4165\pm0.1161}$ & $\underline{\bm{0.6879\pm0.0304}}$ & $0.5260\pm0.1203$ & $\bm{0.7032\pm0.0564}$\\\hline
\textbf{OPF-US1} & $\bm{0.9259\pm0.0251}$ & $\bm{0.9314\pm0.0256}$ & $\bm{0.4040\pm0.1487}$ & $0.6375\pm0.0349$ & $\bm{0.5539\pm0.1298}$ & $\bm{0.6911\pm0.0566}$\\\hline
\textbf{OPF-US2} & $\underline{\bm{0.9319\pm0.0230}}$ & $\underline{\bm{0.9359\pm0.0214}}$ & $\underline{\bm{0.4360\pm0.1303}}$ & $0.6768\pm0.0339$ & $\underline{\bm{0.6088\pm0.1544}}$ & $\underline{\bm{0.7223\pm0.0578}}$\\\hline
\textbf{OPF-US3} & $\bm{0.9275\pm0.0260}$ & $\bm{0.9311\pm0.0269}$ & $0.3457\pm0.1514$ & $0.6190\pm0.0408$ & $\bm{0.5443\pm0.1151}$ & $\bm{0.7173\pm0.0499}$\\\hline\hline
\textbf{CNN~\cite{hart1968condensed}} & $0.8854\pm0.0485$ & $0.8753\pm0.0397$ & $0.3858\pm0.1233$ & $0.6184\pm0.0477$ & $0.3021\pm0.1339$ & $0.6816\pm0.0620$\\\hline
\textbf{RBU~\cite{koziarski2020radial}} & $\bm{0.9185\pm0.0345}$ & $0.9101\pm0.0327$ & $0.3601\pm0.1418$ & $0.5798\pm0.0466$ & $0.4053\pm0.1140$ & $0.6565\pm0.0489$\\\hline
\textbf{NearMiss-1~\cite{maniWLID2003}} & $0.9187\pm0.0266$ & $0.9071\pm0.0271$ & $0.3849\pm0.1099$ & $0.5706\pm0.0444$ & $0.3208\pm0.0935$ & $0.6592\pm0.0462$\\\hline
\textbf{NearMiss-2~\cite{maniWLID2003}} & $0.9182\pm0.0264$ & $0.9086\pm0.0331$ & $0.3742\pm0.1309$ & $0.5773\pm0.0485$ & $0.3531\pm0.0778$ & $0.6734\pm0.0475$\\\hline
\textbf{NearMiss-3~\cite{maniWLID2003}} & $0.9070\pm0.0296$ & $0.9041\pm0.0333$ & $0.3835\pm0.1444$ & $0.5682\pm0.0476$ & $0.1518\pm0.0414$ & $0.6852\pm0.0449$\\

\hhline{-|-|-|-|-|-|-|}
\hhline{-|-|-|-|-|-|-|}
\hhline{-|-|-|-|-|-|-|}
\end{tabular}}
\label{t.results_UndersampleF1_medical}
\end{center}
\end{table*}

Table~\ref{t.results_UndersampleF1_general} highlights that baseline methods performed slightly better over the general-purpose datasets than in the other groups (Tables~\ref{t.results_UndersampleF1_electrical} and~\ref{t.results_UndersampleF1_medical}). In this case, they reached results at least statistically similar to the best ones in two datasets, i.e., Indian Liver and Vertebral Column. However, OPF-based undersampling approaches are again paramount in the context, reaching the highest average F1 values in all datasets, as well as statistically similar results among themselves in most cases.

\begin{table*}[!htb]
\caption{Average F1 score values considering the undersampling tasks over general-purpose datasets.}
\begin{center}
\renewcommand{\arraystretch}{1.0}
\setlength{\tabcolsep}{6pt}
\resizebox{\textwidth}{!}{
\begin{tabular}{c|c|c|c|c|c|c}
\hhline{-|-|-|-|-|-|-|}
\hhline{-|-|-|-|-|-|-|}
\hhline{-|-|-|-|-|-|-|}
{\cellcolor[HTML]{EFEFEF}{\textbf{Technique / Dataset}}}  & {\cellcolor[HTML]{EFEFEF}{\textbf{Indian Liver}}} & {\cellcolor[HTML]{EFEFEF}{\textbf{Secom}}} & {\cellcolor[HTML]{EFEFEF}{\textbf{Seismic Bumps}}} & {\cellcolor[HTML]{EFEFEF}{\textbf{Spam}}} & {\cellcolor[HTML]{EFEFEF}{\textbf{Vertebral Column}}} & {\cellcolor[HTML]{EFEFEF}{\textbf{Wilt}}}\\ \hline
\textbf{OPF-US} & $\underline{\bm{0.5043\pm0.0520}}$ & $\underline{\bm{0.1915\pm0.0472}}$ & $\bm{0.1711\pm0.0435}$ & $0.8691\pm0.0157$ & $\bm{0.6415\pm0.1210}$ & $0.3112\pm0.0363$\\\hline
\textbf{OPF-US1} & $\bm{0.4829\pm0.0668}$ & $0.1532\pm0.0639$ & $\bm{0.1772\pm0.0721}$ & $0.8751\pm0.0158$ & $\bm{0.6553\pm0.1392}$ & $\underline{\bm{0.6495\pm0.0601}}$\\\hline
\textbf{OPF-US2} & $\bm{0.4975\pm0.0492}$ & $\bm{0.1729\pm0.0526}$ & $\underline{\bm{0.1917\pm0.0663}}$ & $0.8680\pm0.0159$ & $\bm{0.6353\pm0.1247}$ & $\bm{0.6424\pm0.0515}$\\\hline
\textbf{OPF-US3} & $0.4258\pm0.0550$ & $0.1027\pm0.0683$ & $0.1454\pm0.0630$ & $\underline{\bm{0.8817\pm0.0140}}$ & $\underline{\bm{0.6664\pm0.1215}}$ & $\bm{0.6416\pm0.0668}$\\\hline\hline
\textbf{CNN~\cite{hart1968condensed}} & $\bm{0.4826\pm0.0668}$ & $0.1651\pm0.0496$ & $0.1246\pm0.0385$ & $0.8196\pm0.0282$ & $\bm{0.6476\pm0.1328}$ & $0.4054\pm0.0846$\\\hline
\textbf{RBU~\cite{koziarski2020radial}} & $0.4599\pm0.0555$ & $0.1253\pm0.0305$ & $0.1270\pm0.0426$ & $0.8558\pm0.0170$ & $\bm{0.6447\pm0.1197}$ & $0.1314\pm0.0445$\\\hline
\textbf{NearMiss-1~\cite{maniWLID2003}} & $0.4077\pm0.0661$ & $0.1716\pm0.0539$ & $0.1451\pm0.0284$ & $0.8511\pm0.0147$ & $\bm{0.6340\pm0.1508}$ & $0.1188\pm0.0168$\\\hline
\textbf{NearMiss-2~\cite{maniWLID2003}} & $0.4762\pm0.0501$ & $0.1512\pm0.0562$ & $0.0945\pm0.0218$ & $0.8480\pm0.0138$ & $0.6195\pm0.1348$ & $0.1127\pm0.0134$\\\hline
\textbf{NearMiss-3~\cite{maniWLID2003}} & $\bm{0.5008\pm0.0610}$ & $0.1265\pm0.0304$ & $0.1064\pm0.0247$ & $0.8186\pm0.0189$ & $\bm{0.6604\pm0.1257}$ & $0.2625\pm0.0438$\\
\hhline{-|-|-|-|-|-|-|}
\hhline{-|-|-|-|-|-|-|}
\hhline{-|-|-|-|-|-|-|}
\end{tabular}}
\label{t.results_UndersampleF1_general}
\end{center}
\end{table*}

\subsection{Best Results and Hybrid Approaches}
\label{ss.undersamplingResults}

This section compares the F1 score values over the original datasets (i.e., without any sample generation) and the best techniques considering both oversampling and undersampling tasks. Moreover, it also compares the results with three proposed hybrid models that employ an undersampling operation followed by an step approach, i.e., OPF-US1-$\text{{O}}^2$PF, OPF-US2-$\text{{O}}^2$PF, and OPF-US3-$\text{{O}}^2$PF. 

Considering the DGA scenario, Table~\ref{t.results_electrical_best} shows that OPF-US3-$\text{{O}}^2$PF provided the best results in general, comprising the most accurate overall values in half of the cases, as well as results statistically similar to the best in all scenarios. Besides, OPF-US3 also performed remarkably well, obtaining the best results according to the Wilcoxon signed-rank test in five datasets, and the most accurate overall results in two. 

\begin{table*}[!htb]
\caption{F1 score values concerning the best approaches from Tables~\ref{t.results_OversampleF1_electrical} and~\ref{t.results_UndersampleF1_electrical} and hybrid versions over the DGA datasets.}
\begin{center}
\renewcommand{\arraystretch}{1.0}
\setlength{\tabcolsep}{6pt}
\resizebox{\textwidth}{!}{
\begin{tabular}{c|c|c|c|c|c|c}
\hhline{-|-|-|-|-|-|-|}
\hhline{-|-|-|-|-|-|-|}
\hhline{-|-|-|-|-|-|-|}
{\cellcolor[HTML]{EFEFEF}{\textbf{Technique / Dataset}}} & {\cellcolor[HTML]{EFEFEF}{\textbf{1069\_5gt}}} & {\cellcolor[HTML]{EFEFEF}{\textbf{1069\_7gt}}} & {\cellcolor[HTML]{EFEFEF}{\textbf{1086\_5ge}}} & {\cellcolor[HTML]{EFEFEF}{\textbf{1086\_7ge}}} & {\cellcolor[HTML]{EFEFEF}{\textbf{1143\_5gte}}} & {\cellcolor[HTML]{EFEFEF}{\textbf{1143\_7gte}}}\\ \hline
\multirow{1}{*}{\textbf{OPF (ORIGINAL)}} & $0.7087\pm0.0915$ & $\bm{0.6804\pm0.1032}$ & $0.7659\pm0.0992$ & $0.5330\pm0.1028$ & $\bm{0.8008\pm0.0603}$ & $0.6697\pm0.0909$\\ \hline\hline
\multirow{1}{*}{\textbf{$\text{{O}}^2$PF}} & $\bm{0.7282\pm0.1116}$ & $\bm{0.6249\pm0.1097}$ & $0.7328\pm0.1052$ & $0.5501\pm0.0950$ & $\bm{0.7981\pm0.0526}$ & $0.6871\pm0.0822$\\ \hline
\multirow{1}{*}{\textbf{$\text{O}^2$PF$_{RI}$}} & $0.7115\pm0.0896$ & $\bm{0.6798\pm0.1005}$ & $0.7485\pm0.1089$ & $0.5453\pm0.1001$ & $\underline{\bm{0.8095\pm0.0570}}$ & $0.6857\pm0.0875$\\ \hline
\multirow{1}{*}{\textbf{$\text{O}^2$PF$_{P}$}} & $\bm{0.7394\pm0.0961}$ & $0.6091\pm0.1249$ & $0.6886\pm0.1366$ & $\bm{0.5629\pm0.1207}$ & $\bm{0.8029\pm0.0752}$ & $0.6918\pm0.0941$\\ \hline
\multirow{1}{*}{\textbf{$\text{O}^2$PF$_{WI}$}} & $0.7038\pm0.1090$ & $0.5925\pm0.0802$ & $0.6956\pm0.1190$ & $\bm{0.5596\pm0.1131}$ & $0.7843\pm0.0770$ & $0.6922\pm0.0885$\\ \hline\hline
\multirow{1}{*}{\textbf{Borderline SMOTE~\cite{han:05borderline}}} & $0.6892\pm0.0794$ & $0.5977\pm0.1151$ & $0.7119\pm0.1213$ & $\bm{0.5698\pm0.1066}$ & $0.7469\pm0.0701$ & $0.6862\pm0.0812$\\ \hline
\multirow{1}{*}{\textbf{SOMO~\cite{douzas2017self}}} & $0.7087\pm0.0915$ & $\bm{0.6830\pm0.1035}$ & $0.7659\pm0.0992$ & $0.5369\pm0.1063$ & $0.7996\pm0.0590$ & $0.6714\pm0.0908$\\ \hline\hline
\multirow{1}{*}{\textbf{OPF-US}} & $0.5250\pm0.0776$ & $0.3593\pm0.0900$ & $0.4846\pm0.0976$ & $0.3023\pm0.0825$ & $0.6645\pm0.0720$ & $0.5184\pm0.0951$\\ \hline
\multirow{1}{*}{\textbf{OPF-US3}} & $\bm{0.7274\pm0.0995}$ & $\underline{\bm{0.6853\pm0.1131}}$ & $\underline{\bm{0.8071\pm0.0883}}$ & $\bm{0.5970\pm0.1073}$ & $\bm{0.7969\pm0.0728}$ & $0.6960\pm0.0964$\\ \hline\hline
\multirow{1}{*}{\textbf{OPF-US1-$\text{{O}}^2$PF}} & $0.7317\pm0.1050$ & $0.5807\pm0.0982$ & $0.7260\pm0.1038$ & $0.5669\pm0.1077$ & $\bm{0.7956\pm0.0639}$ & $0.6895\pm0.0731$\\ \hline
\multirow{1}{*}{\textbf{OPF-US2-$\text{{O}}^2$PF}} & $0.7007\pm0.1084$ & $0.5878\pm0.0945$ & $0.7369\pm0.1061$ & $0.5093\pm0.1021$ & $0.7737\pm0.0800$ & $0.6489\pm0.0885$\\ \hline
\multirow{1}{*}{\textbf{OPF-US3-$\text{{O}}^2$PF}} & $\underline{\bm{0.7718\pm0.0854}}$ & $\bm{0.6673\pm0.1120}$ & $\bm{0.8030\pm0.1025}$ & $\underline{\bm{0.6198\pm0.1010}}$ & $\bm{0.8060\pm0.0715}$ & $\underline{\bm{0.7325\pm0.0743}}$\\
\hhline{-|-|-|-|-|-|-|}
\hhline{-|-|-|-|-|-|-|}
\hhline{-|-|-|-|-|-|-|}
\end{tabular}}
\label{t.results_electrical_best}
\end{center}
\end{table*}

Regarding the medical datasets, Table~\ref{t.results_medical_best} shows that the overwhelming majority of the best results were obtained by the OPF-based undersampling and hybrid approaches, followed by $\text{O}^2$PF variants. Regarding the baselines, the only succeeded case was obtained through SOMO over the Diagnostic dataset. 

\begin{table*}[!htb]
\caption{F1 score values concerning the best approaches from Tables~\ref{t.results_OversampleF1_medical} and~\ref{t.results_UndersampleF1_medical} and hybrid versions over the medical datasets.}
\begin{center}
\renewcommand{\arraystretch}{1.0}
\setlength{\tabcolsep}{6pt}
\resizebox{\textwidth}{!}{
\begin{tabular}{c|c|c|c|c|c|c}
\hhline{-|-|-|-|-|-|-|}
\hhline{-|-|-|-|-|-|-|}
\hhline{-|-|-|-|-|-|-|}
{\cellcolor[HTML]{EFEFEF}{\textbf{Technique / Dataset}}} & {\cellcolor[HTML]{EFEFEF}{\textbf{Diagnostic}}} & {\cellcolor[HTML]{EFEFEF}{\textbf{Diagnostic II}}} & {\cellcolor[HTML]{EFEFEF}{\textbf{Prognostic}}} & {\cellcolor[HTML]{EFEFEF}{\textbf{DDR}}} & {\cellcolor[HTML]{EFEFEF}{\textbf{CC}}} & {\cellcolor[HTML]{EFEFEF}{\textbf{MM}}}\\ \hline
\multirow{1}{*}{\textbf{OPF (ORIGINAL)}} & $0.9290\pm0.0241$ & $0.9086\pm0.0331$ & $0.4136\pm0.1363$ & $0.6140\pm0.0348$ & $0.4900\pm0.1414$ & $0.6359\pm0.0829$\\ \hline\hline
\multirow{1}{*}{\textbf{$\text{{O}}^2$PF}} & $\bm{0.9290\pm0.0227}$ & $0.9240\pm0.0290$ & $0.4034\pm0.1282$ & $0.6125\pm0.0354$ & $\bm{0.5666\pm0.1437}$ & $0.6497\pm0.0524$\\ \hline
\multirow{1}{*}{\textbf{$\text{O}^2$PF$_{RI}$}} & $\bm{0.9321\pm0.0238}$ & $0.9143\pm0.0357$ & $\underline{\bm{0.4589\pm0.1064}}$ & $0.5804\pm0.0489$ & $0.5178\pm0.1362$ & $0.6616\pm0.0596$\\ \hline
\multirow{1}{*}{\textbf{$\text{O}^2$PF$_{WI}$}} & $\bm{0.9324\pm0.0248}$ & $0.9175\pm0.0303$ & $\bm{0.4110\pm0.1186}$ & $0.5804\pm0.0495$ & $\bm{0.5774\pm0.1263}$ & $0.6313\pm0.0819$\\ \hline\hline
\multirow{1}{*}{\textbf{Borderline SMOTE~\cite{han:05borderline}}} & $0.9231\pm0.0278$ & $0.9080\pm0.0353$ & $0.4175\pm0.1266$ & $0.5803\pm0.0460$ & $0.5414\pm0.1109$ & $0.6710\pm0.0479$\\ \hline
\multirow{1}{*}{\textbf{SOMO~\cite{douzas2017self}}} & $\underline{\bm{0.9359\pm0.0246}}$ & $0.9167\pm0.0318$ & $0.4028\pm0.1328$ & $0.5804\pm0.0478$ & $0.5121\pm0.1306$ & $0.6224\pm0.0627$\\ \hline\hline
\multirow{1}{*}{\textbf{OPF-US}} & $\bm{0.9317\pm0.0236}$ & $\bm{0.9334\pm0.0228}$ & $\bm{0.4165\pm0.1161}$ & $\underline{\bm{0.6879\pm0.0304}}$ & $0.5260\pm0.1203$ & $\bm{0.7032\pm0.0564}$\\ \hline
\multirow{1}{*}{\textbf{OPF-US2}} & $\bm{0.9319\pm0.0230}$ & $\bm{0.9359\pm0.0214}$ & $\bm{0.4360\pm0.1303}$ & $0.6768\pm0.0339$ & $\bm{0.6088\pm0.1544}$ & $\bm{0.7223\pm0.0578}$\\ \hline\hline
\multirow{1}{*}{\textbf{OPF-US1-$\text{{O}}^2$PF}} & $\bm{0.9291\pm0.0278}$ & $\underline{\bm{0.9366\pm0.0243}}$ & $\bm{0.4259\pm0.1602}$ & $0.6250\pm0.0392$ & $\bm{0.5852\pm0.1142}$ & $0.6949\pm0.0709$\\ \hline
\multirow{1}{*}{\textbf{OPF-US2-$\text{{O}}^2$PF}} & $\bm{0.9341\pm0.0230}$ & $\bm{0.9357\pm0.0200}$ & $\bm{0.4276\pm0.1360}$ & $0.6569\pm0.0319$ & $\bm{0.5991\pm0.1433}$ & $\underline{\bm{0.7314\pm0.0413}}$\\ \hline
\multirow{1}{*}{\textbf{OPF-US3-$\text{{O}}^2$PF}} & $\bm{0.9304\pm0.0256}$ & $\bm{0.9343\pm0.0255}$ & $0.3799\pm0.1230$ & $0.6164\pm0.0395$ & $\underline{\bm{0.6094\pm0.1012}}$ & $\bm{0.7232\pm0.0515}$\\
\hhline{-|-|-|-|-|-|-|}
\hhline{-|-|-|-|-|-|-|}
\hhline{-|-|-|-|-|-|-|}
\end{tabular}}
\label{t.results_medical_best}
\end{center}
\end{table*}

\begin{sloppypar}
Finally, Table~\ref{t.results_general_best} exhibits the results considering the best oversampling/undersampling approaches, as well as the hybrid models, concerning the general-purpose datasets. Such results confirm the behavior observed above in the other groups of datasets, i.e., better results are provided by OPF-based undersampling and hybrid models, followed by $\text{{O}}^2$PF and its variations. 
\end{sloppypar}

\begin{table*}[!htb]
\caption{F1 score values concerning the best approaches from Tables~\ref{t.results_OversampleF1_general} and~\ref{t.results_UndersampleF1_general} and hybrid versions over the general-purpose datasets.}
\begin{center}
\renewcommand{\arraystretch}{1.0}
\setlength{\tabcolsep}{6pt}
\resizebox{\textwidth}{!}{
\begin{tabular}{c|c|c|c|c|c|c}
\hhline{-|-|-|-|-|-|-|}
\hhline{-|-|-|-|-|-|-|}
\hhline{-|-|-|-|-|-|-|}
{\cellcolor[HTML]{EFEFEF}{\textbf{Technique / Dataset}}} & {\cellcolor[HTML]{EFEFEF}{\textbf{Indian Liver}}} & {\cellcolor[HTML]{EFEFEF}{\textbf{Secom}}} & {\cellcolor[HTML]{EFEFEF}{\textbf{Seismic Bumps}}} & {\cellcolor[HTML]{EFEFEF}{\textbf{Spam}}} & {\cellcolor[HTML]{EFEFEF}{\textbf{Vertebral Column}}} & {\cellcolor[HTML]{EFEFEF}{\textbf{Wilt}}}\\ \hline
\multirow{1}{*}{\textbf{OPF (ORIGINAL)}} & $0.4532\pm0.0709$ & $0.1533\pm0.0645$ & $0.1611\pm0.0806$ & $0.8683\pm0.0144$ & $\bm{0.6557\pm0.1388}$ & $\bm{0.6411\pm0.0658}$\\ \hline\hline
\multirow{1}{*}{\textbf{$\text{{O}}^2$PF}} & $\bm{0.4850\pm0.0566}$ & $0.1536\pm0.0357$ & $0.1970\pm0.0877$ & $0.8712\pm0.0144$ & $\bm{0.6651\pm0.1290}$ & $0.5189\pm0.0507$\\ \hline
\multirow{1}{*}{\textbf{$\text{O}^2$PF$_{MI}$}} & $\bm{0.4972\pm0.0615}$ & $0.1572\pm0.0366$ & $0.1987\pm0.0808$ & $0.8740\pm0.0132$ & $\bm{0.6524\pm0.1333}$ & $0.6162\pm0.0466$\\ \hline
\multirow{1}{*}{\textbf{$\text{O}^2$PF$_{P}$}} & $\bm{0.4791\pm0.0731}$ & $\bm{0.1664\pm0.0696}$ & $0.1913\pm0.0779$ & $0.8721\pm0.0128$ & $\bm{0.6497\pm0.1400}$ & $0.5661\pm0.0891$\\ \hline\hline
\multirow{1}{*}{\textbf{SMOTE~\cite{ChawlaJAIR:02}}} & $\bm{0.4699\pm0.0652}$ & $0.1577\pm0.0358$ & $0.1581\pm0.0588$ & $0.8706\pm0.0155$ & $\bm{0.6703\pm0.1276}$ & $\underline{\bm{0.6677\pm0.0498}}$\\ \hline
\multirow{1}{*}{\textbf{MWMOTE~\cite{Barua2014MWMOTE}}} & $0.4700\pm0.0739$ & $0.1501\pm0.0381$ & $0.1754\pm0.0650$ & $0.8728\pm0.0183$ & $\underline{\bm{0.6751\pm0.1217}}$ & $\bm{0.6593\pm0.0541}$\\ \hline\hline
\multirow{1}{*}{\textbf{OPF-US}} & $\underline{\bm{0.5043\pm0.0520}}$ & $\underline{\bm{0.1915\pm0.0472}}$ & $0.1711\pm0.0435$ & $0.8691\pm0.0157$ & $0.6415\pm0.1210$ & $0.3112\pm0.0363$\\ \hline
\multirow{1}{*}{\textbf{OPF-US1}} & $\bm{0.4829\pm0.0668}$ & $0.1532\pm0.0639$ & $0.1772\pm0.0721$ & $0.8751\pm0.0158$ & $\bm{0.6553\pm0.1392}$ & $\bm{0.6495\pm0.0601}$\\ \hline
\multirow{1}{*}{\textbf{OPF-US2}} & $\bm{0.4975\pm0.0492}$ & $\bm{0.1729\pm0.0526}$ & $\bm{0.1917\pm0.0663}$ & $0.8680\pm0.0159$ & $0.6353\pm0.1247$ & $0.6424\pm0.0515$\\ \hline
\multirow{1}{*}{\textbf{OPF-US3}} & $0.4258\pm0.0550$ & $0.1027\pm0.0683$ & $0.1454\pm0.0630$ & $0.8817\pm0.0140$ & $\bm{0.6664\pm0.1215}$ & $\bm{0.6416\pm0.0668}$\\ \hline\hline
\multirow{1}{*}{\textbf{OPF-US1-$\text{{O}}^2$PF}} & $\bm{0.5032\pm0.0556}$ & $0.1528\pm0.0386$ & $0.1985\pm0.0889$ & $0.8775\pm0.0171$ & $\bm{0.6579\pm0.1328}$ & $0.5299\pm0.0467$\\ \hline
\multirow{1}{*}{\textbf{OPF-US2-$\text{{O}}^2$PF}} & $\bm{0.4989\pm0.0489}$ & $0.1605\pm0.0442$ & $\bm{0.2152\pm0.0694}$ & $0.8707\pm0.0130$ & $0.6330\pm0.1207$ & $0.4929\pm0.0538$\\ \hline
\multirow{1}{*}{\textbf{OPF-US3-$\text{{O}}^2$PF}} & $\bm{0.4807\pm0.0528}$ & $\bm{0.1541\pm0.0459}$ & $\underline{\bm{0.2297\pm0.0770}}$ & $\underline{\bm{0.8848\pm0.0133}}$ & $\bm{0.6718\pm0.1219}$ & $0.5387\pm0.0422$\\
\hhline{-|-|-|-|-|-|-|}
\hhline{-|-|-|-|-|-|-|}
\hhline{-|-|-|-|-|-|-|}
\end{tabular}}
\label{t.results_general_best}
\end{center}
\end{table*}

\subsection{Discussion}
\label{ss.discussion}

The experiments conducted in the paper, which comprised a total of $18$ datasets and divided into three groups, i.e., DGA, medical, and general-purpose, confirm the efficiency and robustness of the proposed OPF-based models for oversampling, compared against seven state-of-the-art methods, and undersampling tasks, compared against five state-of-the-art approaches, as well as the three hybrid approaches. 

In general, $\text{{O}}^2$PF and its variants outperformed the baselines considering the oversampling methods in most cases, stressing that the general performance of standard $\text{{O}}^2$PF and some variations performed better than others according to each group of datasets. Concerning the undersampling tasks, OPF-US and its variations obtained promising results, outperforming all state-of-the-art approaches in virtually any scenario. 

An additional round of experiments was conducted in the previous section, where the best oversampling and undersampling techniques are compared against the OPF-based hybrid models. One can observe that both undersampling and hybrid approaches compete for the best results, leaving little space for oversampling or the baseline techniques. 

We understand that creating samples is more complicated than just removing them in imbalanced scenarios. Also, we are not eliminating any sample but the less relevant ones according to a ranked list of scores. On the other hand, oversampling methods may change the distribution of data when the synthetic samples are generated, ending up in a more complicated classification task.

Regarding the hybrid models, they were statistically similar to undersampling approaches in most cases. However, for some datasets, the hybrid models outperformed by far the second-best techniques. We can highlight some results over DGA datasets in Table~\ref{t.results_electrical_best}, e.g., 1069\_5gt, in which OPF-US3-O$^2$OPF obtained an F1 score of $0.77$, and the second-best one achieved $0.73$. Similar behaviors can be observed in 1143\_7gte and Seismic Bumps datasets as well.

Such results lead us to believe that oversampling is essential, but we first need to sanitize data, e.g., counterfeit samples, to avoid their propagation towards the synthetic samples. Since OPF explores the strength of connectivity among samples in the feature spaces no matter they are distant or not, it can learn data distribution better than others in some situations.

\subsection{Visual Analysis of Data Distribution}
\label{ss.resultsBreast}

\begin{figure*}[!htb]
  \centerline{
    \begin{tabular}{ccc}
	\raisebox{-.5\height}{\includegraphics[width=.28\textwidth]{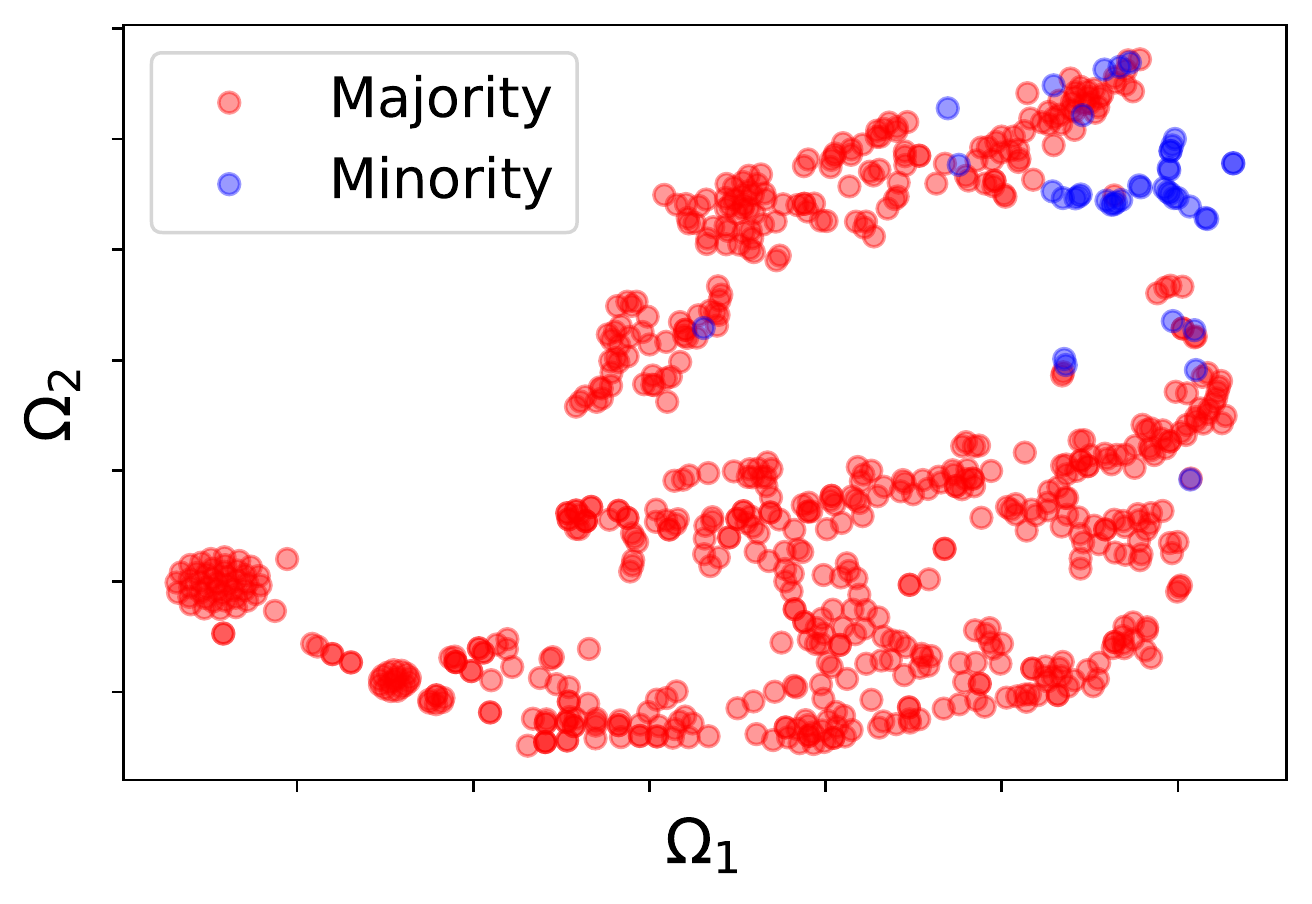}} &
	\raisebox{-.5\height}{\includegraphics[width=.28\textwidth]{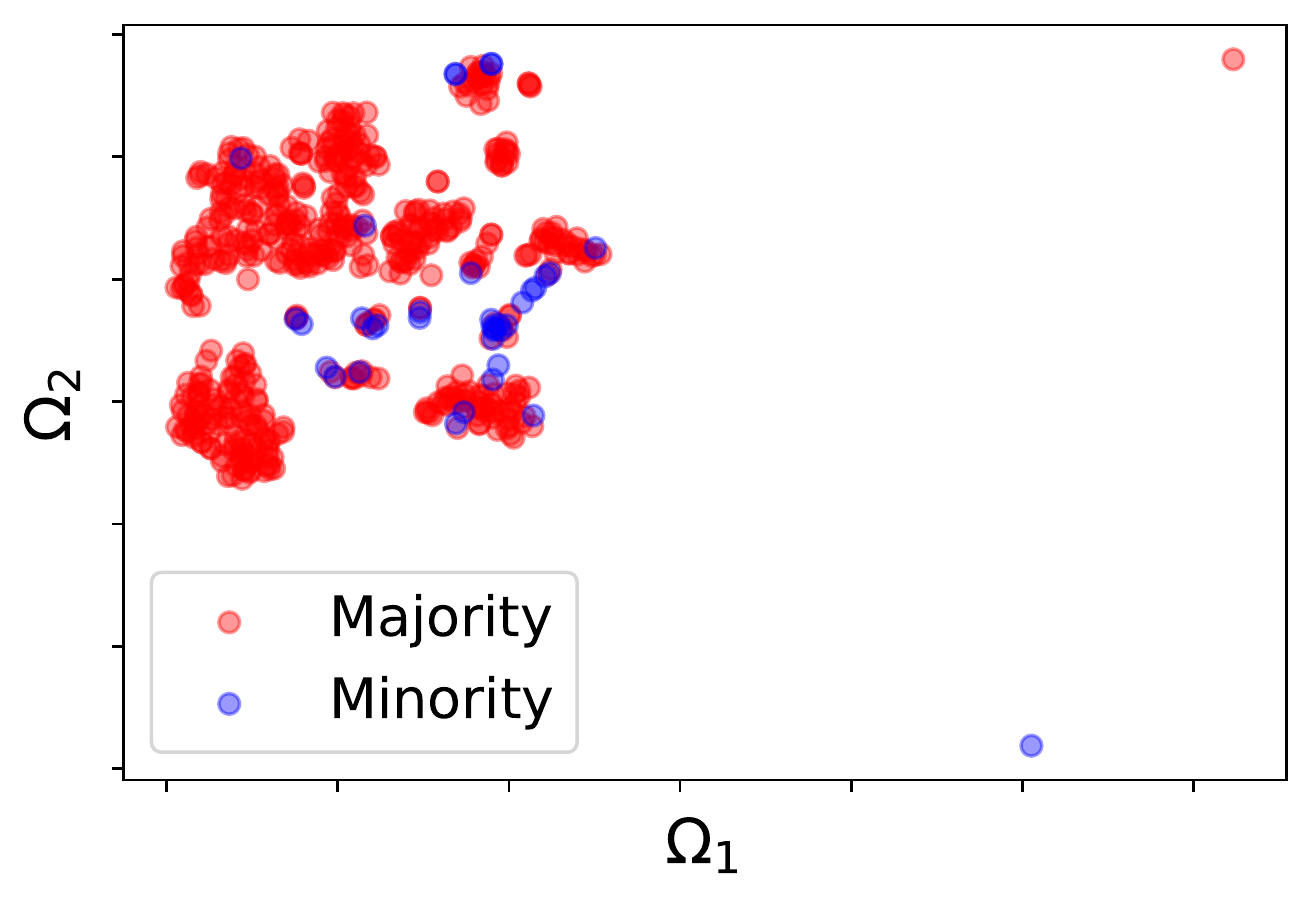}} &
	\raisebox{-.5\height}{\includegraphics[width=.28\textwidth]{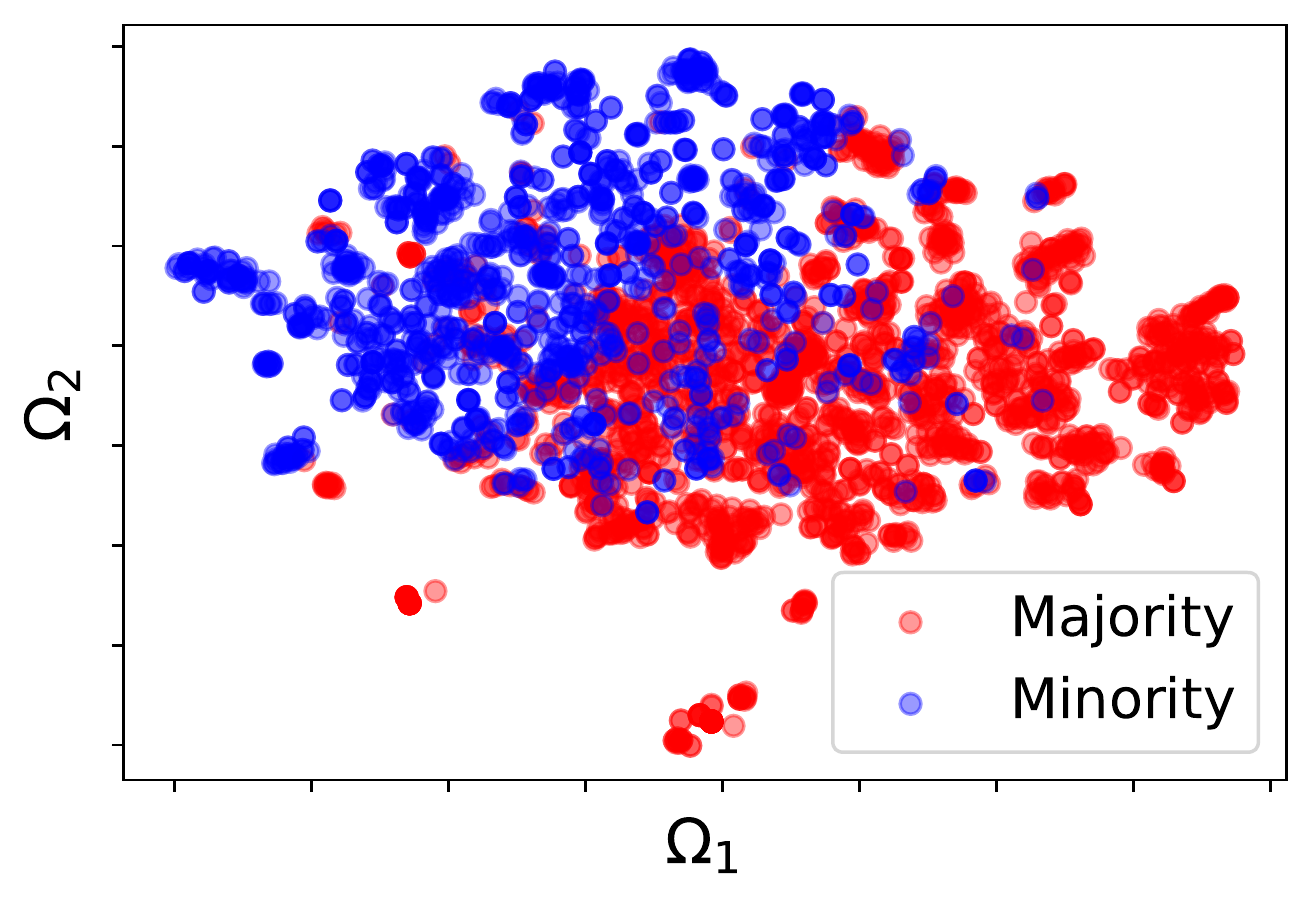}}  \\
	(a) & (b) & (c) 
    \end{tabular}}
  \centerline{
    \begin{tabular}{ccc}
	\raisebox{-.5\height}{\includegraphics[width=.28\textwidth]{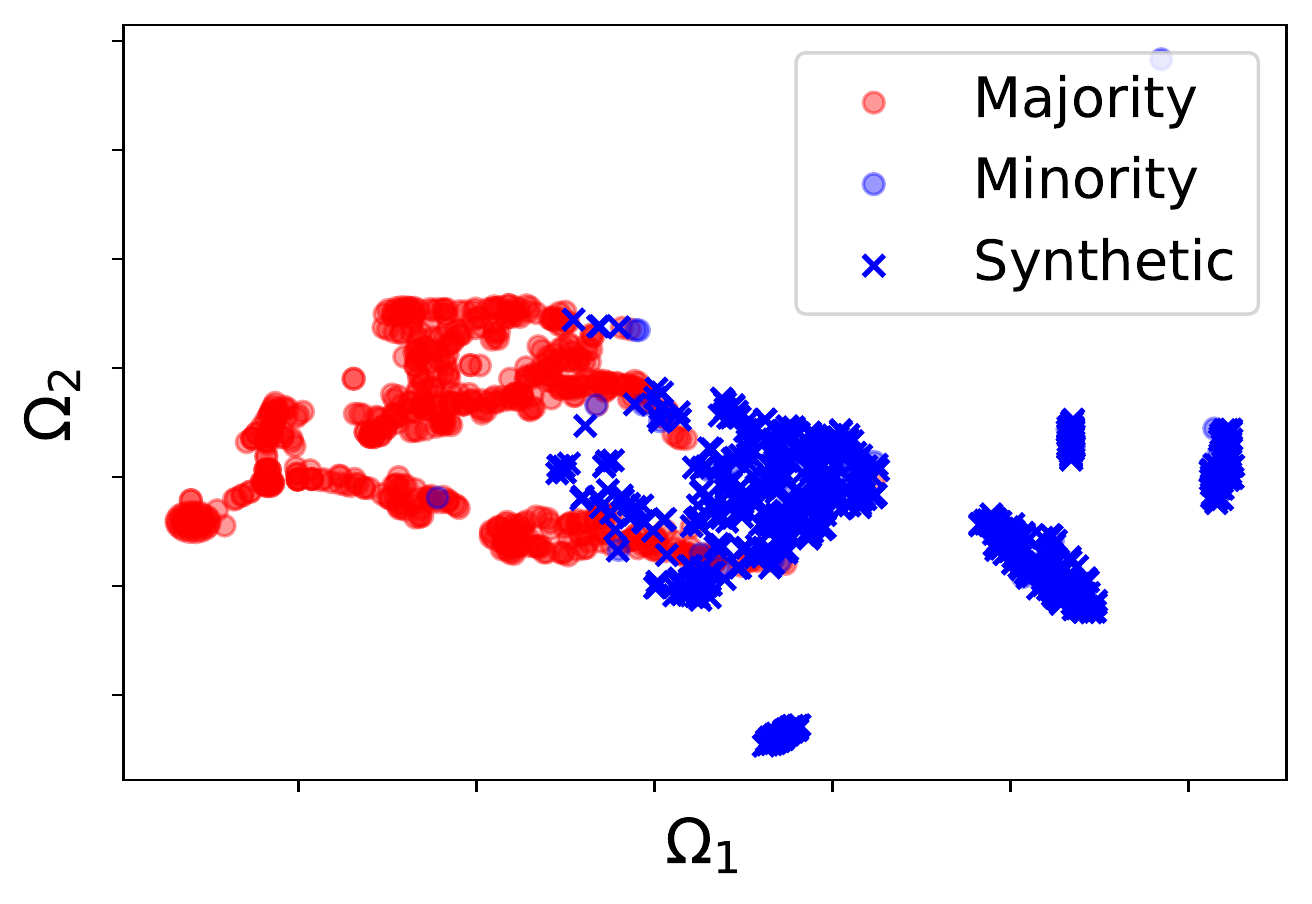}}  &
	\raisebox{-.5\height}{\includegraphics[width=.28\textwidth]{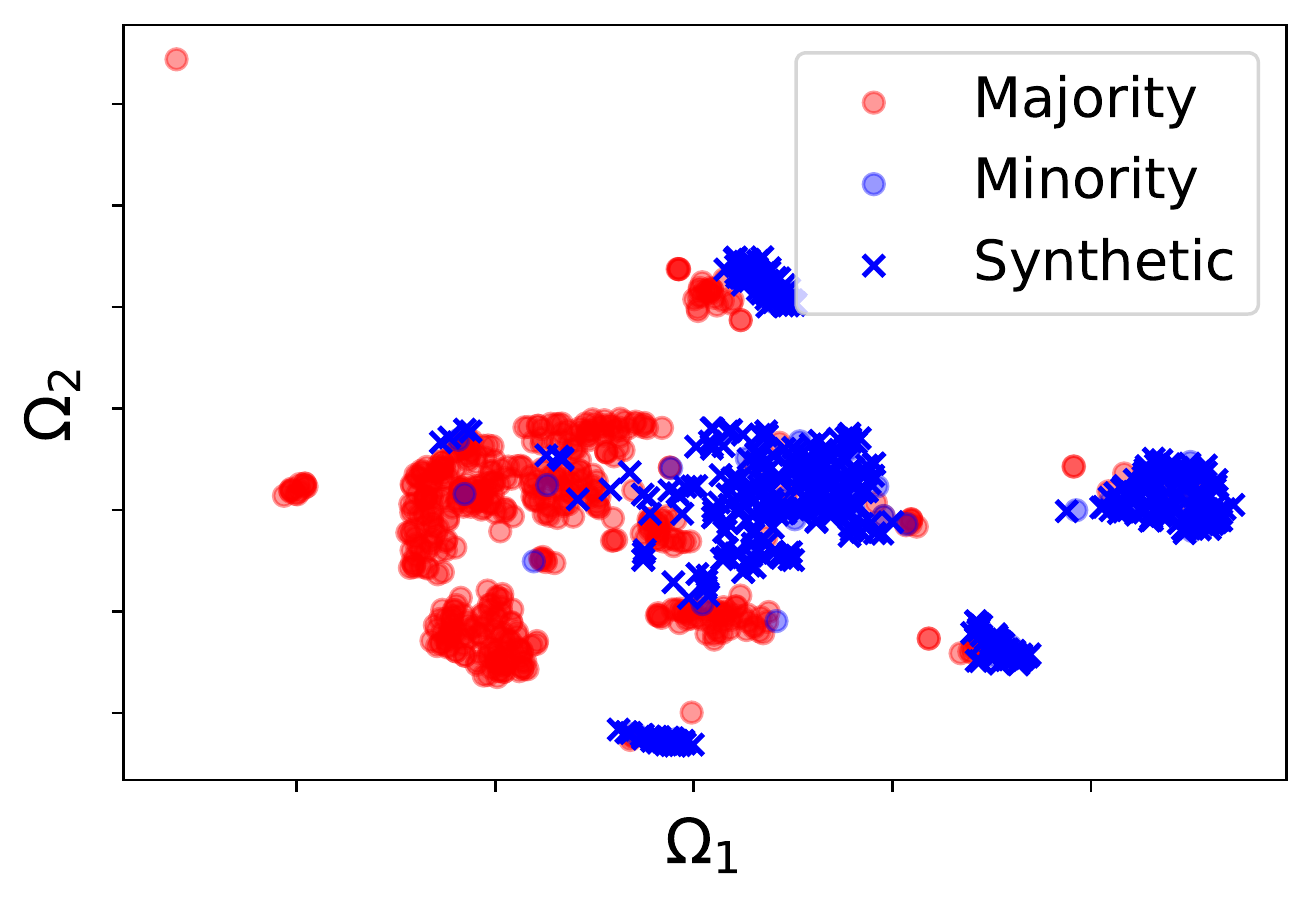}}  &
	\raisebox{-.5\height}{\includegraphics[width=.28\textwidth]{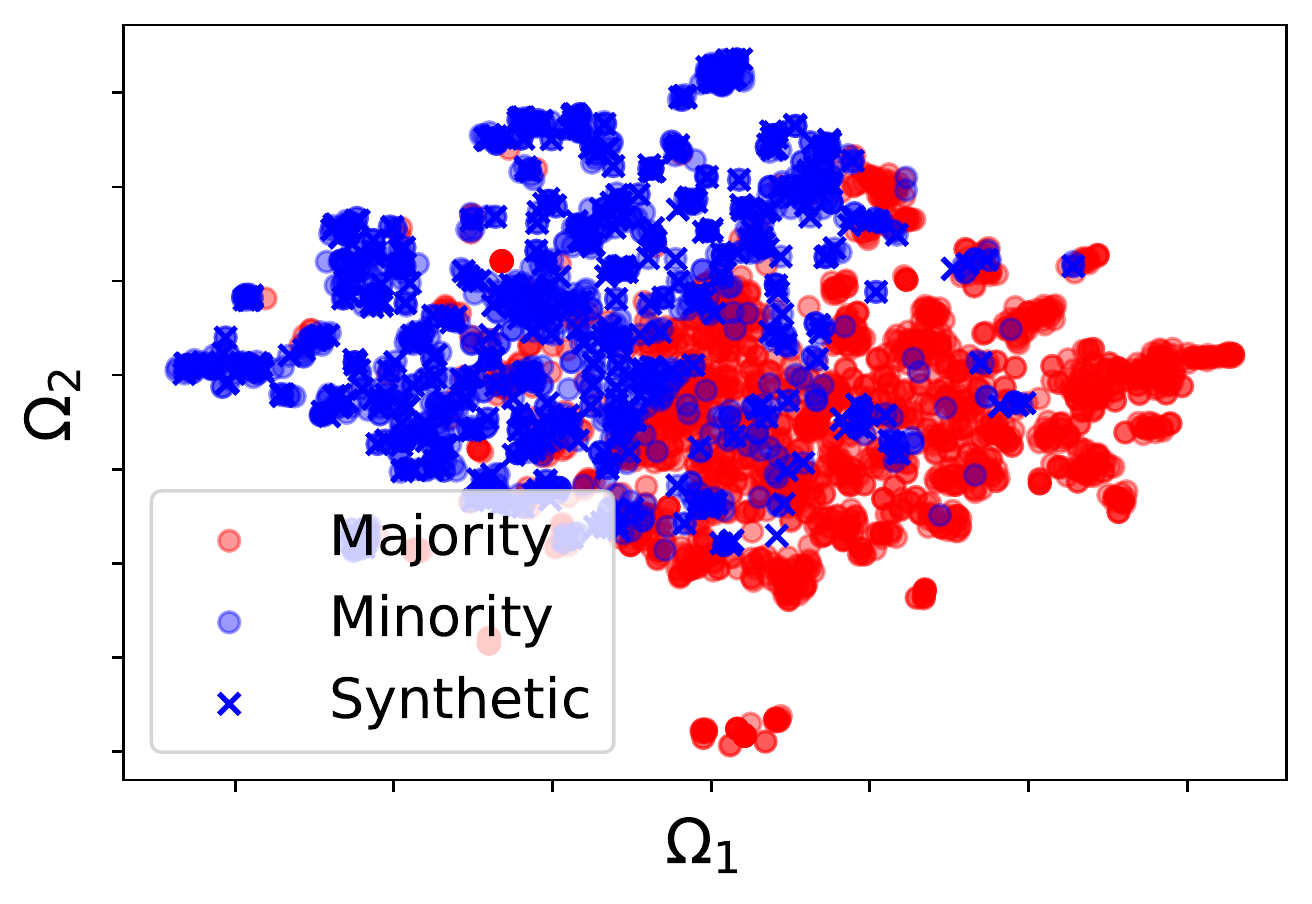}}\\
	(d)  & (e) & (f) 
    \end{tabular}}	
  \centerline{
    \begin{tabular}{ccc}
	\raisebox{-.5\height}{\includegraphics[width=.28\textwidth]{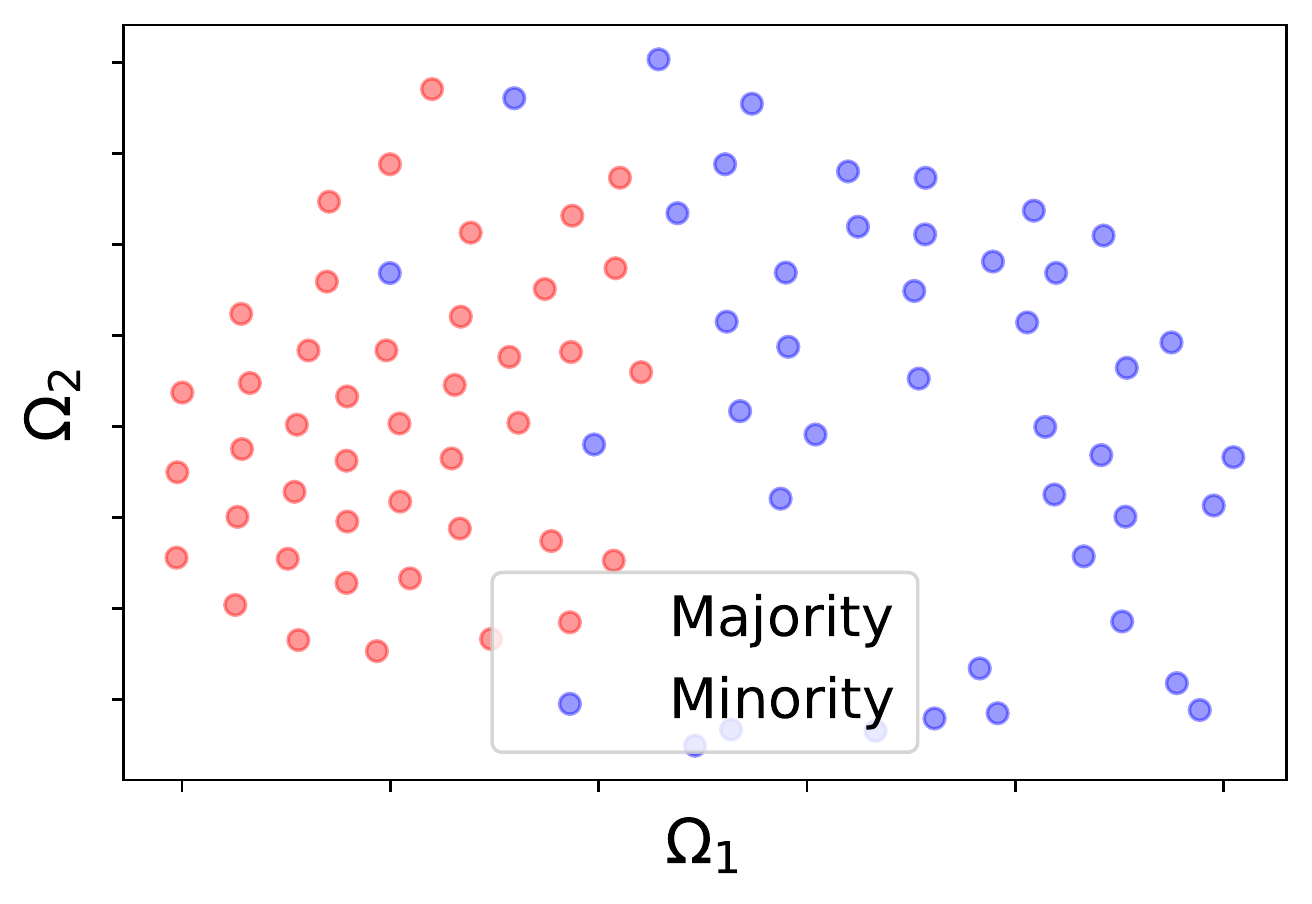}}  &
	\raisebox{-.5\height}{\includegraphics[width=.28\textwidth]{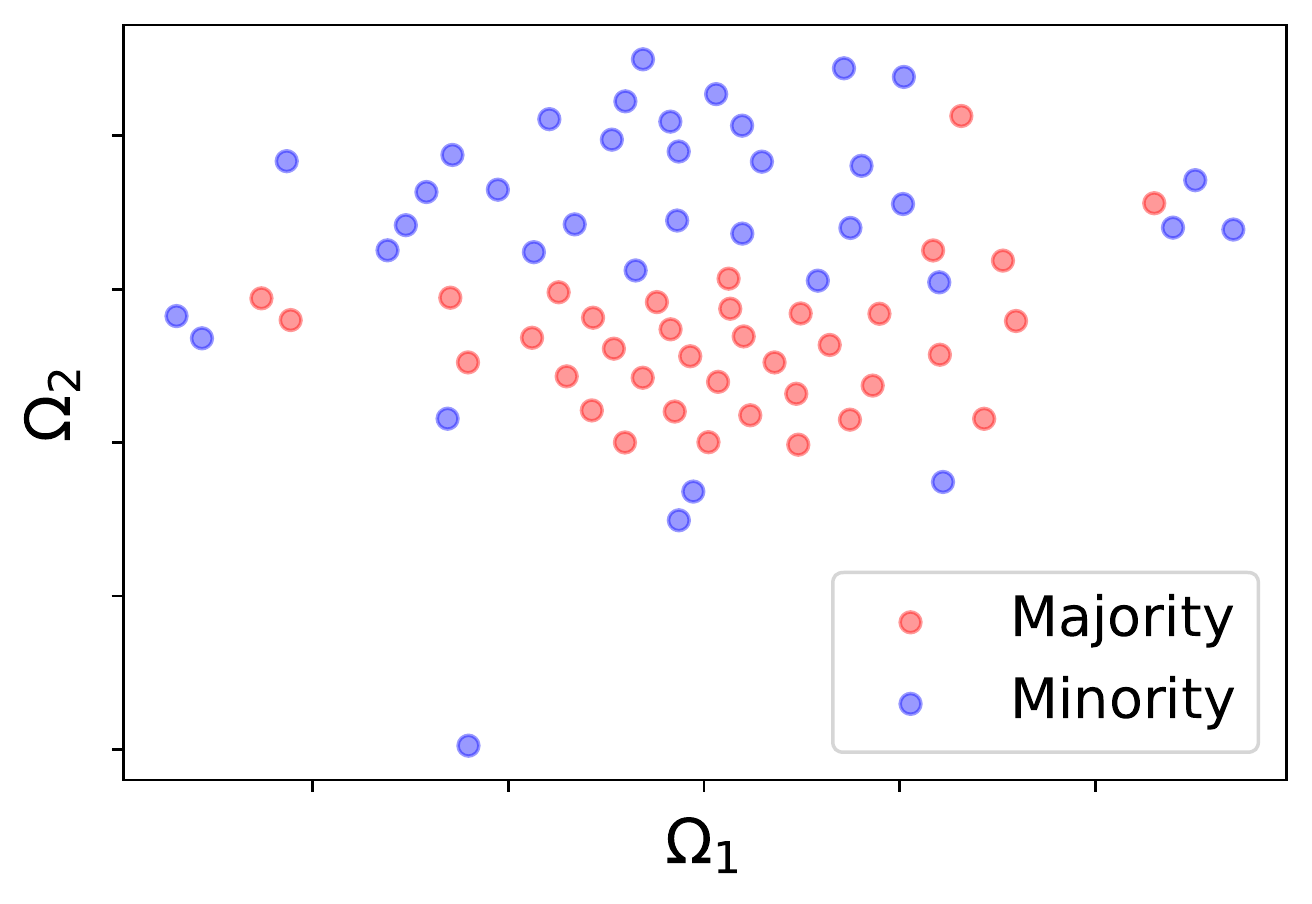}}  &
	\raisebox{-.5\height}{\includegraphics[width=.28\textwidth]{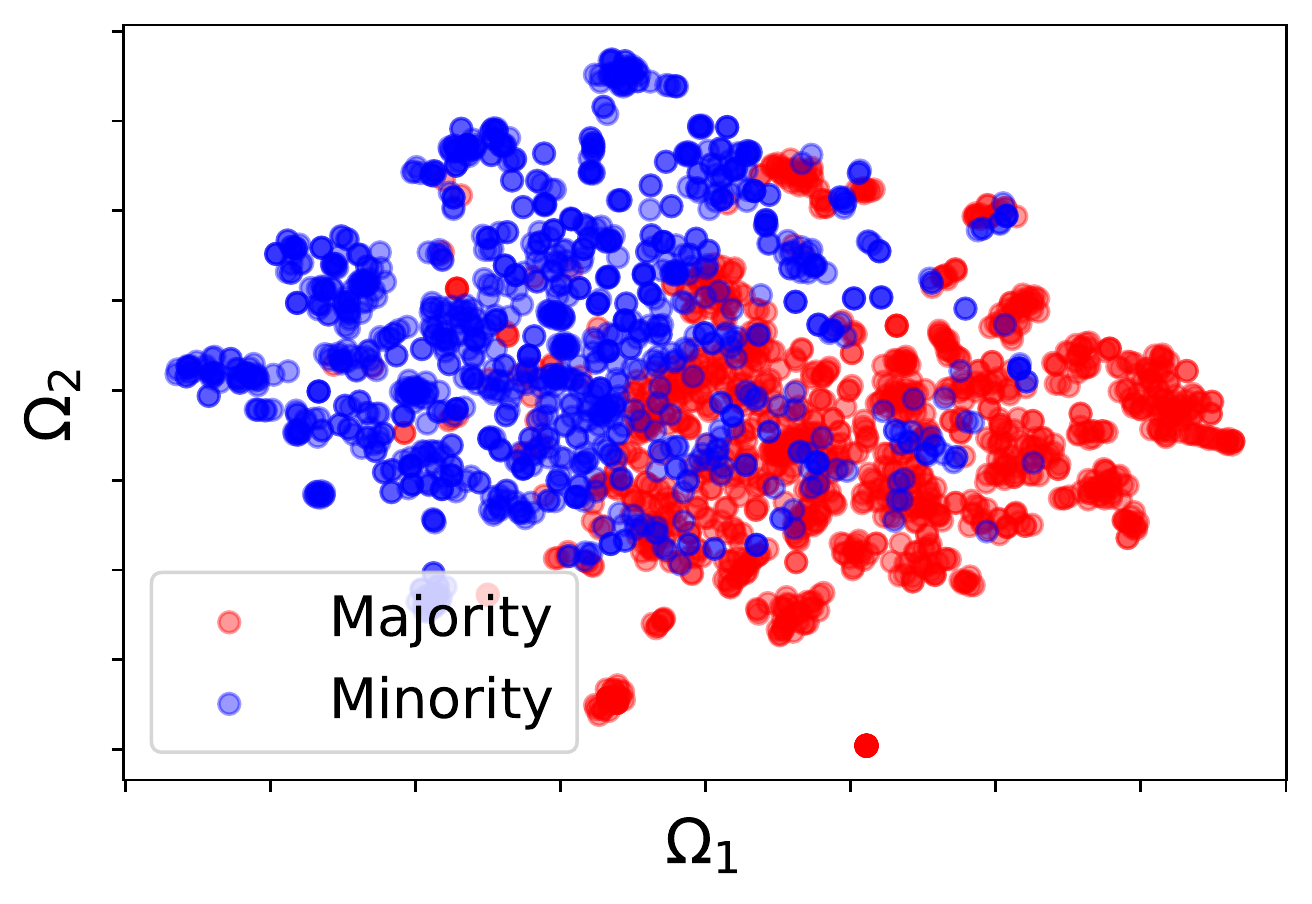}}\\
	(g)  & (h) & (i) 
    \end{tabular}}	
  \centerline{
    \begin{tabular}{ccc}
	\raisebox{-.5\height}{\includegraphics[width=.28\textwidth]{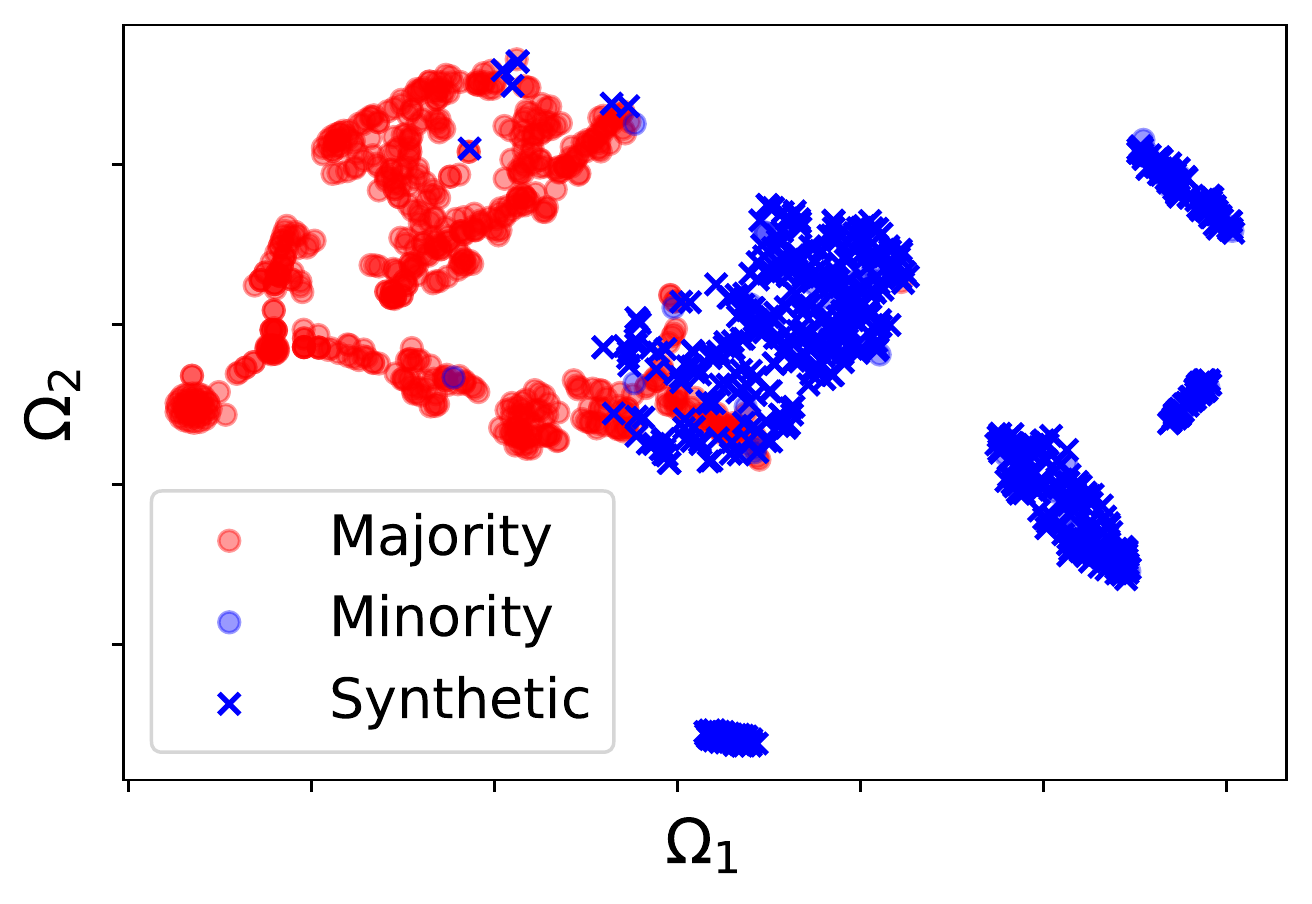}}  &
	\raisebox{-.5\height}{\includegraphics[width=.28\textwidth]{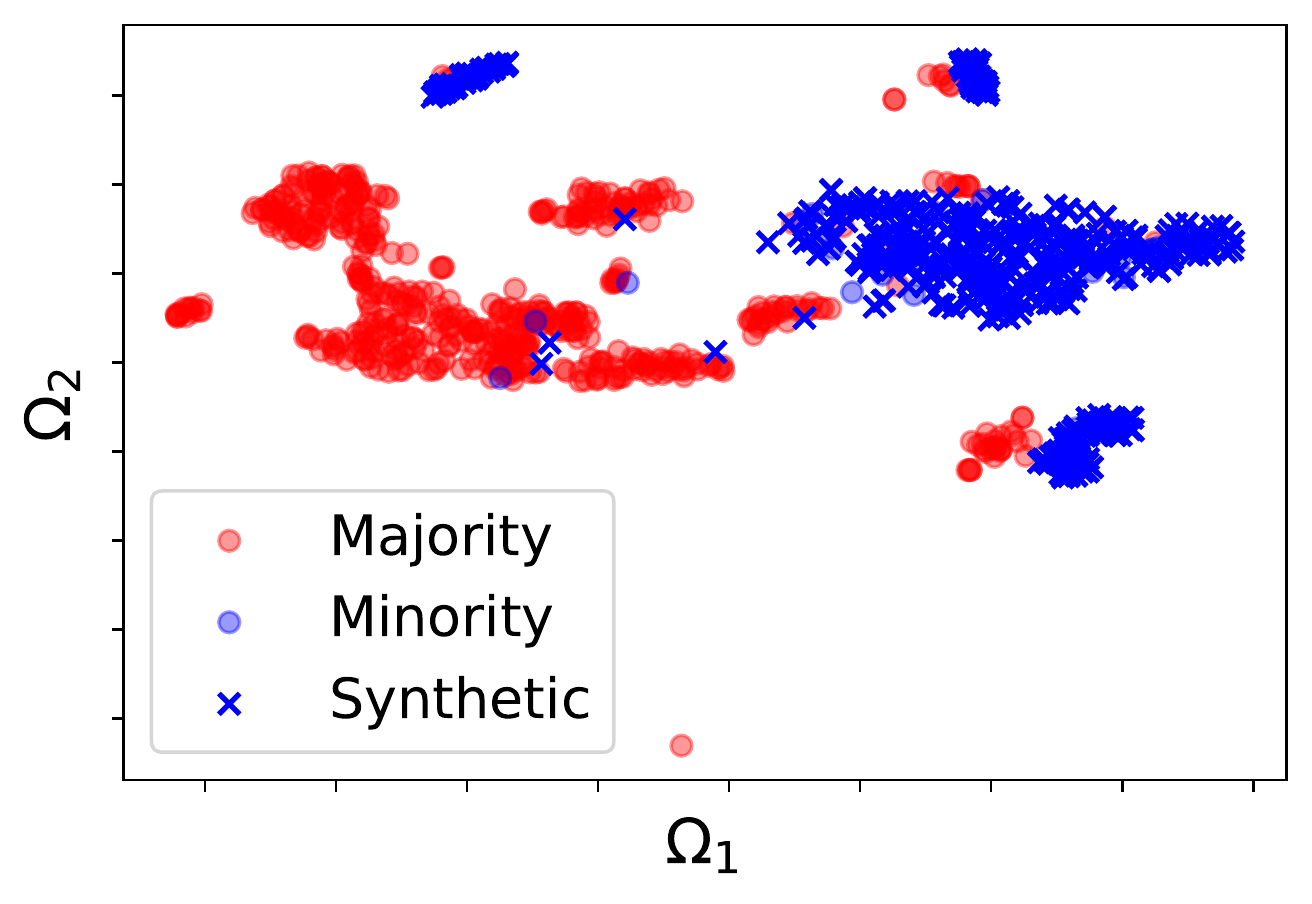}}  &
	\raisebox{-.5\height}{\includegraphics[width=.28\textwidth]{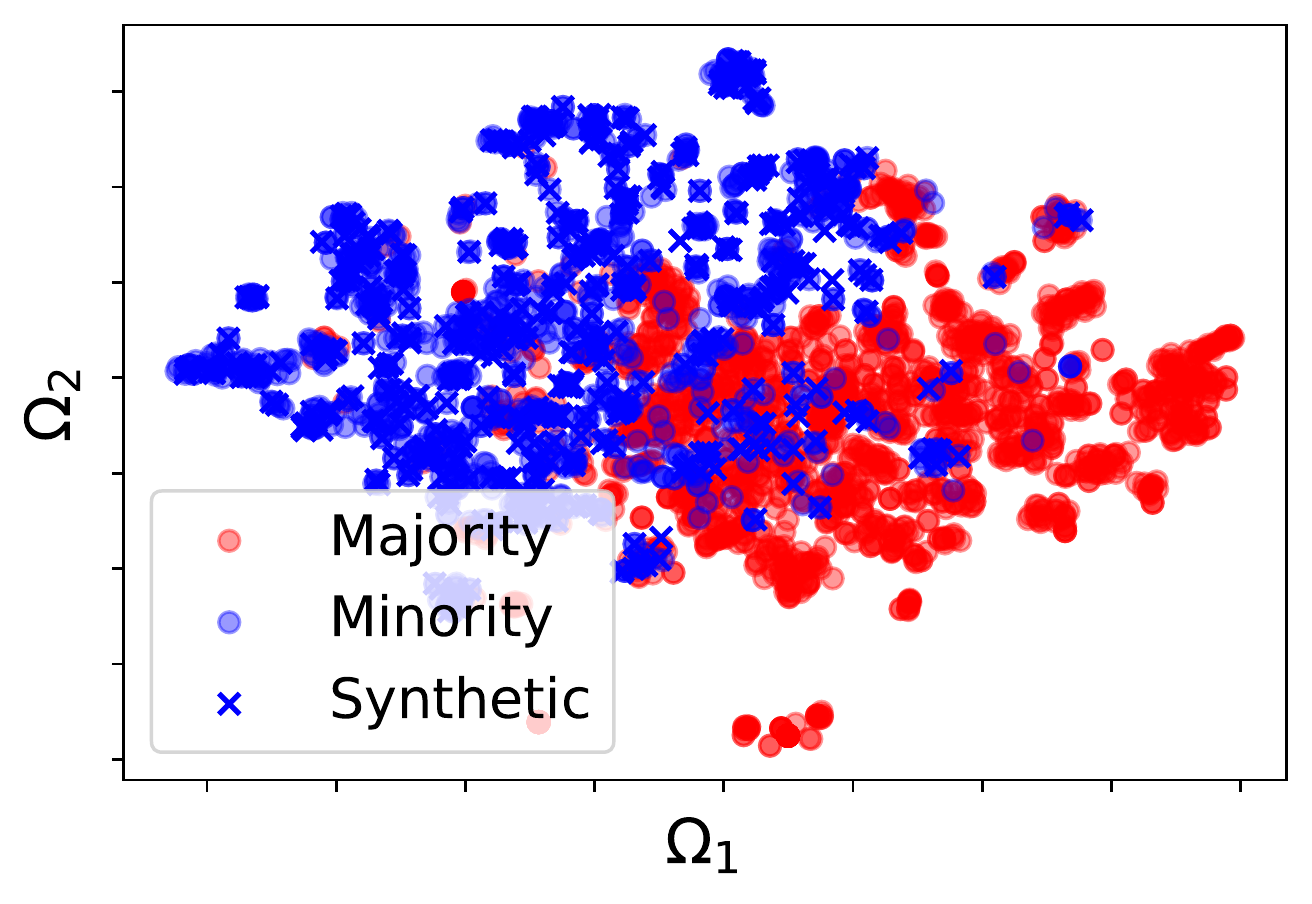}}\\
	(j)  & (k) & (l) \\
    \end{tabular}}	
    \caption{The leftmost column denotes the 1069\_5gt dataset, the middle one stands for the Cervical Cancer dataset, and the rightmost column represents the Spam dataset. Further, the first row denotes the original datasets, the second row represents oversampling through $\text{O}^2$PF, undersampling through OPF-US is depicted in the third row, and undersampling followed by oversampling using OPF-US3-$\text{O}^2$PF is illustrated in the fourth row.}
  \label{f.plots}
\end{figure*}

In this section, we showed the feature space provided by the proposed approaches is well behaved, ending up in better recognition rates. Figure~\ref{f.plots} illustrates this behavior for three datasets, i.e., one per group of applications, where the leftmost column denotes the 1069\_5gt dataset, in the middle one can report to the Cervical Cancer dataset, and the rightmost column stands for the Spam dataset. For the sake of visualization, we applied the well-known t-SNE~\cite{maaten2008visualizing} approach for 2-D data transformation. Moreover, the rows correspond to different scenarios: original dataset in the first row, oversampling through $\text{O}^2$PF in the second row, undersampling through OPF-US in the third row, and  undersampling followed by oversampling using OPF-US3-$\text{O}^2$PF in the fourth row.

Considering 1069\_5gt dataset, despite the label imbalance and some minority-class outliers (Figure~\ref{f.plots}a), it presents a reasonably well-behaved feature space. Such an effect is maximized by the oversampling approach in Figure~\ref{f.plots}d, thus providing more accurate results. On the other hand, pruning majority-class samples do not seem to improve the classification performance in this case (see Figure~\ref{f.plots}g and OPF-US results in Table~\ref{t.results_electrical_best}), since minority-class outliers became more significant in the reduced set of samples. However, pruning the minority-class outliers (see OPF-US3 results in Table~\ref{t.results_electrical_best}) seems to reverse this situation. Moreover, applying the OPF-US3 followed by an oversampling approach (Figure~\ref{f.plots}j and  and OPF-US3-$\text{{O}}^2$PF results in Table~\ref{t.results_electrical_best}) stands for the ideal scenario, obtaining the best overall results.

Cervical Cancer shows itself as a dataset highly imbalanced whose original samples from both classes may look overlapped (Figure~\ref{f.plots}b). However, the pruned version depicted in Figure~\ref{f.plots}h seems to promote a distribution were data separability is straightforward. Further, the hybrid approach benefits from such a phenomenon, thus obtaining the most accurate overall results, as observed in Tables~\ref{t.results_UndersampleF1_medical} and~\ref{t.results_medical_best}.

Although the imbalance presented in the Spam dataset is less incisive, it presents some similarities with 1069\_5gt regarding minority-class outliers among majority class samples, as observed in Figure~\ref{f.plots}c. However, a hybrid approach, where the less significative samples from majority class are pruned and several samples from minority class are properly created, appears to be the best approach for the problem, as presented in Table~\ref{t.results_general_best}.

\section{Conclusions}
\label{s.conclusions}

This paper tackled the problem of handling imbalanced datasets proposing several Optimum-Path Forest-based techniques for both data oversampling and undersampling. More specifically, the paper proposes four $\text{O}^2$PF-based oversampling variations, i.e., $\text{O}^2$PF$_{RI}$, $\text{O}^2$PF$_{MI}$, $\text{O}^2$PF$_{P}$, and $\text{O}^2$PF$_{WI}$, as well as four undersampling approaches, i.e., OPF-US and its variations OPF-US1, OPF-US2, and OPF-US3. Moreover, we also proposed three hybrid models, which combine standard $\text{O}^2$PF with OPF-US1, OPF-US2, and OPF-US3. 

An in-depth section of experiments conducted over $18$ datasets composed of distinct imbalance degrees, and considering $7$ state-of-the-art approaches for oversampling and $5$ other techniques for undersampling purposes, validates the effectiveness of the proposed methods, stressing the undersampling and hybrid methods, which overwhelmingly outperformed the baselines.

Regarding future works, we plan to extend the techniques to deal with multiple labels. Besides, we also intend to consider other variations of the supervised OPF classifier.

\section*{Acknowledgments}
\begin{sloppypar}
The authors are grateful to FAPESP grants \#2020/12101-0, \#2013/07375-0, \#2014/12236-1, \#2017/02286-0, \#2018/21934-5, \#2019/07665-4, and \#2019/18287-0, as well as CNPq grants \#307066/2017-7, and \#427968/2018-6.
\end{sloppypar}

\section*{References}

\bibliography{references}

\end{document}